\documentclass[sigconf]{acmart}

\usepackage{booktabs} 
\usepackage{url}  
\usepackage{graphicx}
\graphicspath{{figs/}}
\usepackage{amssymb}
\usepackage{bm}
\usepackage{amsmath}
\usepackage{enumitem}
\usepackage{float}
\usepackage{bbm}
\usepackage[many]{tcolorbox}
\usetikzlibrary{shadows}
\usepackage{natbib}
\usepackage{subcaption}
\setcopyright{rightsretained}
\usepackage{xcolor}
\usepackage{colortbl}
\usepackage{amsthm}

\usepackage[utf8]{inputenc}
\usepackage[T1]{fontenc}

\newtheorem{theorem}{Theorem}

\newtheorem{manualtheoreminner}{Theorem}
\newenvironment{manualtheorem}[1]{%
  \renewcommand\themanualtheoreminner{#1}%
  \manualtheoreminner
}{\endmanualtheoreminner}

\copyrightyear{2020} 
\acmYear{2020} 
\setcopyright{acmcopyright}
\acmConference[ACM CHIL '20]{ACM Conference on Health, Inference, and Learning}{April 2--4, 2020}{Toronto, ON, Canada}
\acmBooktitle{ACM Conference on Health, Inference, and Learning (ACM CHIL '20), April 2--4, 2020, Toronto, ON, Canada}
\acmPrice{15.00}
\acmDOI{10.1145/3368555.3384456}
\acmISBN{978-1-4503-7046-2/20/04}

\newcommand{\dsopioid}{\textsc{\textbf{Opioid}}}
\newcommand{\dsihdp}{\textsc{\textbf{IHDP}}}
\newcommand{\dssyn}{\textsc{\textbf{Synthetic}}}
\newcommand{\method}{\textsc{HEMM}}

\newcommand{\diag}{\mathop{\mathrm{diag}}}

\newcommand\given[1][]{\:#1\vert\:}

\newcommand\independent{\protect\mathpalette{\protect\independenT}{\perp}}
\def\independenT#1#2{\mathrel{\rlap{$#1#2$}\mkern2mu{#1#2}}}

\begin{document}
\title{Interpretable Subgroup Discovery in Treatment Effect Estimation with Application to Opioid Prescribing Guidelines}

\author{Chirag Nagpal,$^{1,2}$ Dennis Wei,$^1$ Bhanukiran Vinzamuri,$^1$ Monica Shekhar,$^3$\\ \textbf{\normalfont Sara E. Berger,$^1$ Subhro Das,$^1$ and Kush R. Varshney$^1$}\\
$^1$IBM Research\\
$^2$Carnegie Mellon University\\
$^3$IBM Global Business Services
}

\renewcommand{\shortauthors}{Nagpal et. al.}
\renewcommand{\shorttitle}{Interpretable Subgroup Discovery in Treatment Effect Estimation}

\begin{abstract}
The dearth of prescribing guidelines for physicians is one key driver of the current opioid epidemic in the United States.  In this work, we analyze medical and pharmaceutical claims data to draw insights on characteristics of patients who are more prone to adverse outcomes after an initial synthetic opioid prescription.  Toward this end, we propose a generative model that allows discovery from observational data of subgroups that demonstrate an enhanced or diminished causal effect due to treatment. Our approach models these sub-populations as a mixture distribution, using sparsity to enhance interpretability, while jointly learning nonlinear predictors of the potential outcomes to better adjust for confounding. The approach leads to human-interpretable insights on discovered subgroups, improving the practical utility for decision support.
\end{abstract}

%
%



\maketitle

\section{Introduction}

The United States is in the midst of an opioid addiction epidemic. According to estimates by the Centers for Disease Control and Prevention (CDC), 42,000 people died from opioid overdoses in 2016 and 49,000 in 2017. Overdose mortalities specifically from \emph{synthetic} opioids, such as Fentanyl, have increased exponentially since 1999.\footnote{\url{https://www.cdc.gov/drugoverdose/data/analysis.html}} A major cause of this epidemic is overprescription of opioids (for legitimate pain management) by physicians who lack proper prescribing guidelines \citep{califf2016proactive,makary2017overprescribing}.

One actionable insight for prescribers is characteristics of patients for whom treatment with synthetic opioids, as opposed to natural or semi-synthetic opioids, causes a greater risk of adverse outcomes such as long-term use and addiction than for the general population.  Toward this end, we study causal treatment effect estimation from observational data under heterogeneity, i.e.\ the phenomenon of different individuals having different responses to the same treatment.  In particular, we focus on the discovery of subgroups of patients (really portions of a feature space) that have enhanced or diminished treatment effects.  We aim for the discovered subgroups to be human-interpretable so that the results can be directly used in prescribing guidelines.

\begin{figure}
\centering
\includegraphics[width=0.8\linewidth]{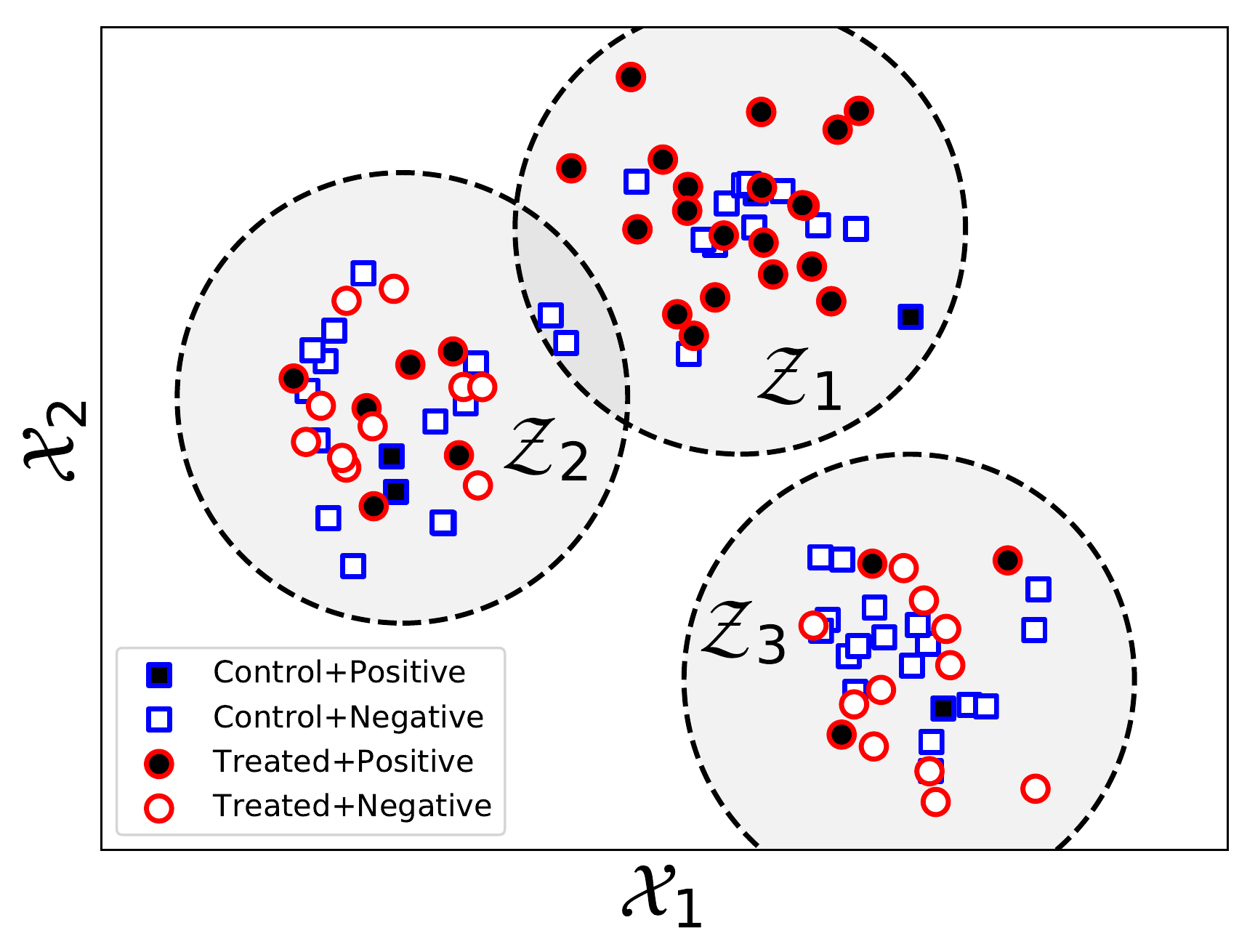}

\caption{The heterogeneous effect subgroup discovery problem. Almost all instances receiving treatment in $\mathcal{Z}_1$ have a positive outcome, while very few in $\mathcal{Z}_3$ do. We are interested in recovering such latent subgroups.}
\label{fig:example}

\end{figure}

We make use of the MarketScan database of medical claims and pharmaceutical claims.  Claims are a form of administrative data that are commonly repurposed for medical studies because they capture the diagnoses, procedures, and prescriptions of patients longitudinally. Inference of causal relationships using observational data is challenging since counterfactual outcomes are not observed and the treated and untreated populations may have underlying differences that affect the outcome. However, modern machine learning techniques provide an avenue to overcome the challenges. 

To identify subgroups with different treatment effects, we hypothesize that a latent variable determines the treatment effect of each individual. Moreover, individuals with similar characteristics belong to the same latent subgroup, resulting in similar responses to treatment across the subgroup. Figure \ref{fig:example} is an abstract representation of such a phenomenon.

We propose a Bayesian network to model these subgroups, specifically as a mixture model, along with their corresponding treatment effects. 
Sparsity is induced in the learned mixture component parameters to improve interpretability.  Our approach, which we name the heterogeneous effect mixture model (\method{}), is similar in spirit to causal rule sets for identifying subgroups with enhanced treatment effect \citep{wang2017causal} but does not require hard partitions or assignments.  Moreover, we incorporate nonlinear as well as linear outcome models, 
which increases the expressiveness of the model to better adjust for confounding without sacrificing the interpretability of the subgroup definitions.  We thus benefit from both the interpretability of sparse mixture models and the representation learning capability of neural networks.  In contrast, recent works \cite{louizos2017causal,shalit2016estimating,alaa2017bayesian} that use neural networks or nonparametric methods to estimate heterogeneous treatment effects do not identify subgroups of individuals with similar responses.  While our motivating application is 
opioid use, the proposed approach applies to 
any problem domain requiring the discovery of subgroups 
with heterogeneous responses to actions. 
In this spirit, we also validate our method on synthetic data and the Infant Health and Development Program (IHDP) dataset in terms of its heterogeneous effect estimation and subgroup identification performance.

With respect to opioids, we provide domain expert interpretation of the enhanced treatment effect subgroup discovered using MarketScan data, i.e.~patients at higher risk of adverse outcomes after an initial synthetic opioid prescription.  Some characteristics of this subgroup are well-known and/or reflected in CDC opioid prescribing guidelines \cite{dowell2016CDC}: chronic pain conditions, psychological comorbidities, heart disease and obesity.  The presence of minor injuries and dental/oral conditions in the subgroup can be explained by the common practice of prescribing opioids for post-surgical or intense acute pain.  Lastly, some discovered conditions are unexpected, such as skin infections, abscesses, and reproductive disorders. 

Overall, our contributions can be summarized as follows:
\begin{enumerate}[label=\roman*)]
\item We propose the \method{} for discovering subgroups with enhanced and diminished treatment effects in a potential outcomes causal inference framework, using sparsity to enhance interpretability. 
\item We extend the \method{}'s outcome model to include neural networks to better adjust for confounding and develop a joint inference procedure for the overall graphical model and the neural networks.
\item We demonstrate strong performance in estimating heterogeneous effects and identifying subgroups compared to existing approaches.
\item We apply the methodology to a large-scale medical claims dataset and discover actionable patient subgroups at enhanced risk of adverse outcomes with synthetic opioids.
\end{enumerate}

\section{Related Work}

There is a rich literature of data-oriented research on understanding the patterns and risks of opioid prescribing and addiction in the fields of medicine, medical informatics, and machine learning; some is specifically intended to inform prescribing guidelines.  \citet{Kim2016} conduct a randomized controlled trial and \citet{NeillH2018} analyze spatiotemporal overdose event data, but a large part of the literature works with medical claims data \citep{Parente2004,Edlund2007,Jena2014,zhangexploring,Brat2018,Klueh2018} and similar administrative data \citep{che2017deep} as we do.  However, unlike us, none of these works focus on heterogeneous treatment effects.



The identification of subgroups with heterogeneous or enhanced treatment effects has been addressed in the statistics literature by building separate factual and counterfactual outcome models and then regressing the difference of the two using another 
method, e.g.~a decision tree \citep{su2009subgroup}. This final model can then be deployed to identify subgroups. 
Within this category of approaches, \citet{lipkovich2011subgroup} propose the subgroup identification based on differential effect search (SIDES) algorithm, \citet{Dusseldorp2014} propose the qualitative interaction trees (QUINT) algorithm, and \citet{foster2011subgroup} propose the virtual twins (VT) method. We consider empirical comparisons to these algorithms in the sequel.

\citet{wang2017causal} propose causal rule sets for discovering subgroups with enhanced treatment effect. This is the closest to and an inspiration for our work.  That work seeks to learn discrete human-interpretable rules predictive of enhanced treatment effect and involves optimization by Monte Carlo methods. We consider instead a mixture of experts approach with soft assignment to groups that retains most of the interpretability but allows greater expressiveness and can be optimized via gradient methods. 
Our outcome model \eqref{eqn:outcome}, \eqref{eqn:outcomeCont} also differs from that of \cite{wang2017causal}.  Most importantly, we allow nonlinearity in the form of neural networks whereas \cite{wang2017causal} considers only linear models.  Our model also has a single term representing the main effect of treatment whereas \cite{wang2017causal} has three such terms: a population average, a subgroup term that is always active, and a subgroup term that is only active under treatment.  

Recent papers have proposed estimating heterogeneous/individual treatment effects using neural networks \citep{louizos2017causal,shalit2016estimating} or a Bayesian nonparametric method involving Gaussian processes \cite{alaa2017bayesian}. These methods rely on constructing distributional representations of the factual and counterfactual outcomes that are similar in a statistical sense. While these methods perform well on estimating heterogenous effects, they
 do not identify subgroups of individuals with similar treatment effects and characteristics and are thus less interpretable. This makes the application of such methods to inform policy decisions more difficult.




\section{Heterogeneous Effect Mixture Model}

In this section, we propose a generative mixture model for heterogeneous treatment effects.  
One way to model heterogeneity, and the one in our proposal, is as a finite mixture of components with a different treatment effect model in each component (some enhanced and some diminished).  
For tractability, we keep the form of the mixtures to be the simplest possible: Gaussian-distributed for continuous covariates and Bernoulli-distributed for discrete covariates.  
We encode a preference for components to involve few covariates through a Laplace prior or a group $\ell_{1,2}$ prior on the means of the covariates, described in Section~\ref{sec:sparsity}.  Section~\ref{sec:outcome} presents a model for the outcomes as they depend on treatment, covariates, and mixture membership, including nonlinear dependence on the covariates.


\subsection{Preliminaries}
We adopt the Neyman-Rubin potential outcomes framework \citep{rubin1974estimating} for causal inference. Define random variables $\mathbf{X} \in \mathbb{R}^d$ representing covariates 
and $T \in \{0, 1\}$ as the treatment indicator. The subset of continuous-valued covariates are denoted $\mathbf{X}_{\text{cont}}$ and the discrete covariates (binary-valued or binarized) are denoted $\mathbf{X}_{\text{disc}}$. We will sometimes refer to $T=1$ as `the treatment' and $T=0$ as `the control.'  Corresponding to the levels of treatment are two potential outcomes $Y(0)$ and $Y(1)$, which are the outcomes under $T=0$ and $T=1$ respectively.  These outcomes can be discrete- or continuous-valued.  We are given an observational dataset of samples $\mathcal{D} = \{ (\mathbf{x}_i, t_i, y_i )\}_{i=1}^{N}$ in which only one of the outcomes is observed for each individual: if $t_i = 0$ then $y_i = y(0)_i$, and if $t_i = 1$ then $y_i = y(1)_i$.  

Our interest lies in estimating the conditional average treatment effect (CATE) conditioned on $\mathbf{X}$, defined as 
\[
\tau(\mathbf{X}) = \mathbb{E}\bigl[Y(1) - Y(0) \given \mathbf{X}\bigr].
\]
In this work, the dependence on $\mathbf{X}$ is mediated primarily through subgroup membership, i.e.~members of the same subgroup have similar treatment effects. 
We make the standard assumptions that allow CATE to be identifiable from observational data, namely exchangeability conditioned on the available covariates, $T \perp (Y(0), Y(1)) \given \mathbf{X}$, positivity of the treatment propensity, $0 < p(T=1\given \mathbf{x}) < 1$ for all $\mathbf{x}$, and no dependence between individuals i.e.~the stable unit treatment value assumption (SUTVA) \citep{Rubin1986,HernanR2018}. The first two assumptions are collectively known as strong ignorability (SITA).

For the mixture model proposed in this paper, we additionally define the latent random variable $Z \in \mathcal{Z} = \{1,\ldots,K\}$ to indicate mixture membership. Both the distribution of covariates and the treatment effect are dependent on $Z$ as described next.

\subsection{Generative Model}
\label{sec:genModel}

\begin{figure}
\centering
  \includegraphics[width=0.6\linewidth, trim={4cm 3.0cm 19.0cm 1cm}, clip]{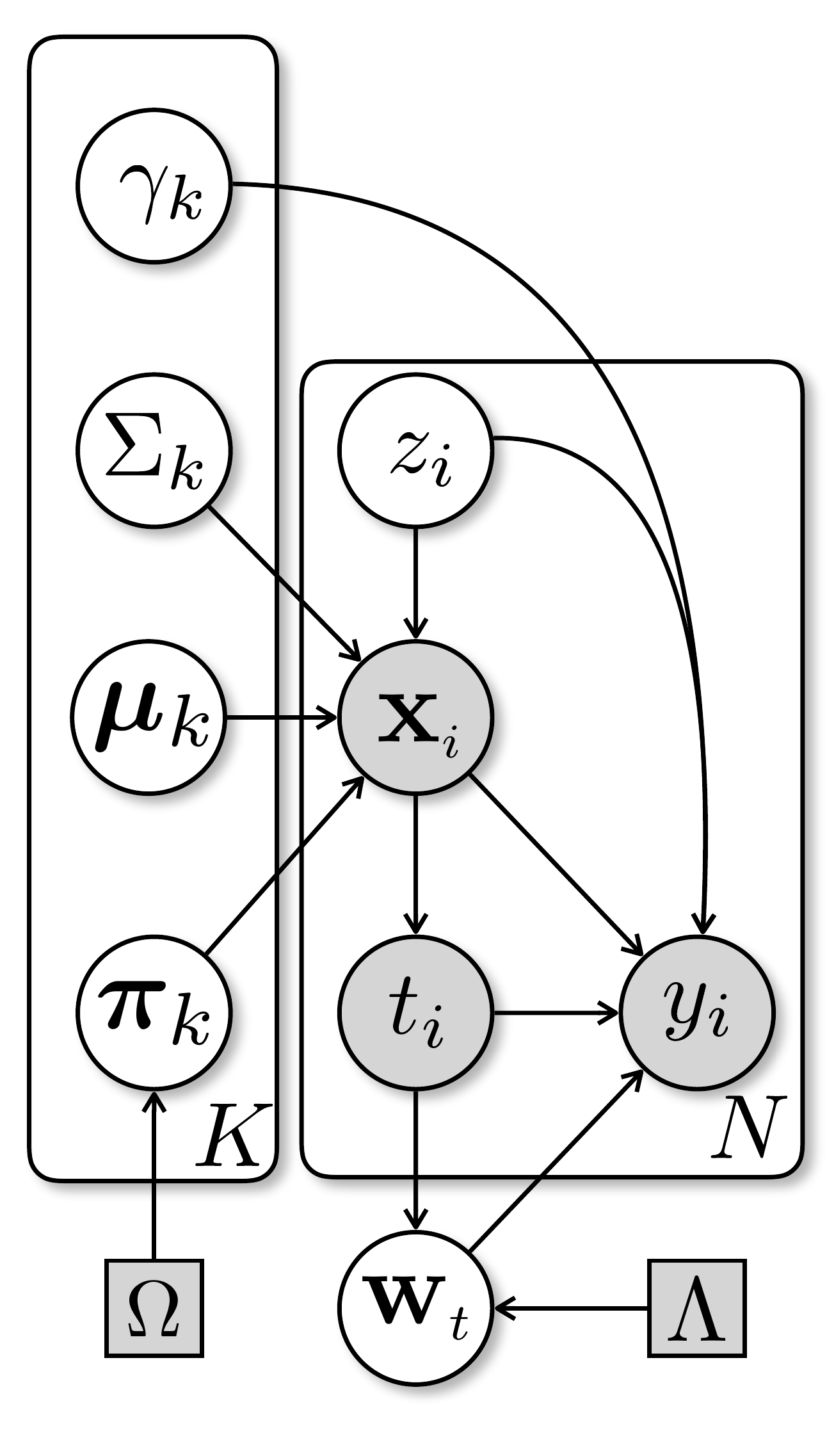}%
\caption{The proposed heterogeneous effect mixture model (\method{}) in plate notation. For each instance $i$, $(\mathbf{x}_i, t_i, y_i)$ are the observed variables and $z_i$ is a latent variable that determines membership in one of the $K$ mixture components. Each component has an associated coefficient $\gamma_k$ that determines the main treatment effect.}
    \label{fig:plate}
\end{figure}

The generative model is presented in Figure \ref{fig:plate} in plate notation. We first give an overview of the distributions and then go into more detail regarding $\mathbf{X}$ and $Y$ in Sections~\ref{sec:sparsity} and \ref{sec:outcome}. 

\begin{enumerate}
\item We draw a sample $z_i$ independently for each individual $i$ that determines the latent group membership. The prior distribution for $Z$ is uniform over the $K$ groups, 
\begin{equation}\label{eqn:Z}
Z \sim \text{Uniform}(K).
\end{equation}
\item Conditioned on the latent group assignment $z_i = k$, 
\begin{enumerate}
\item The $\mathbf{x}_{\text{cont},i}$ are drawn i.i.d.~from a Gaussian distribution with mean $\boldsymbol{\mu}_k$ and covariance $\Sigma_k$:
\begin{equation}\label{eqn:Xcont}
\mathbf{X}_{\text{cont}} \mid z_i=k \sim \text{Normal}(\boldsymbol{\mu}_k, \Sigma_k).
\end{equation}
In this paper, we constrain the off-diagonal elements of $\Sigma_k$ to be $0$ to reduce the number of parameters, although non-diagonal covariances can be easily accommodated.
\item The $\mathbf{x}_{\text{disc},i}$ are drawn i.i.d.~from a multivariate Bernoulli distribution with mean $\boldsymbol{\pi}_k$:
\begin{equation}\label{eqn:Xdisc}
\mathbf{X}_{\text{disc}} \mid z_i = k \sim \text{Bernoulli}(\boldsymbol{\pi}_k).
\end{equation}
We enforce sparsity in $\boldsymbol{\pi}_k$ in order to improve interpretability. We describe this in detail in Section~\ref{sec:sparsity}.
\end{enumerate}

\item Conditioned on the covariates $\mathbf{x}_i$, the treatment assignment $t_i$ is drawn from a Bernoulli distribution whose mean is a function of $\mathbf{x}_i$:
\[
T \given \mathbf{x} \sim \text{Bernoulli}(\rho(\mathbf{x})).
\]
This corresponds to a model for treatment \emph{propensity}. Note from Figure~\ref{fig:plate} that the generative model assumes that $T$ is conditionally independent of $Z$ given $\mathbf{X}$.  Under this assumption, it will be seen in Section~\ref{sec:inference} that inference for the propensity model can be done independently from the other components of the generative model.

\item  Finally, an outcome sample $y_i$ is drawn from a distribution whose mean $\mu_y$ is a function 
of the covariates $\mathbf{x}_i$, treatment assignment $t_i$, and latent group assignment $z_i$.  If $Y$ is binary-valued, the distribution is Bernoulli, 
\begin{displaymath} Y \mid \mathbf{x}, t, z \sim \text{Bernoulli}\bigl(\mu_y(\mathbf{x}, t, z)\bigr),\end{displaymath}
whereas if $Y$ is continuous, the distribution is Gaussian, 
\begin{displaymath} Y \mid \mathbf{x}, t, z \sim \text{Normal}\bigl(\mu_y(\mathbf{x}, t, z), \sigma_y^2 \bigr),\end{displaymath}
where $\sigma_y^2$ is the variance.
The outcome model is discussed further in Section~\ref{sec:outcome}.
\end{enumerate}


Note we are interested in estimating the causal quantity,  $$\mathbb{E}[Y(t)|X]= \mathbb{E}[Y|\textbf{do}(T=t), X] = p(Y|\textbf{do}(T=t), X).$$

Here, the first equality is from definition of interventional quantities and the second equality holds due to $Y$ being binary. 

\begin{theorem}[Identifiability]
\label{thm:identifiability}
Under the Directed Acyclic Graph in Figure. \ref{fig:plate}, $$p(Y|\textbf{do}(T=t), X) = \int_Z p(Y|X,Z,T=t) p(Z|X).$$ 
\end{theorem}
Theorem \ref{thm:identifiability} confirms that we can estimate the CATE from the observational quantities introduced above. The proof is deferred to the Appendix \ref{appx:identifiability}.

\subsection{Sparse Mixture Components for Interpretability} \label{sec:sparsity}
Without further measures, the mixture component means $\boldsymbol{\mu}_k$ and $\boldsymbol{\pi}_k$ learned from data may be dense, making them difficult for a domain expert to interpret.  We hypothesize that a large number of learned mean parameters may have small values and that promoting sparsity through appropriate prior distributions can 
overcome this problem. 
To this end, we experiment with two different sparsity-promoting priors on the means $\boldsymbol{\pi}_k$ of discrete covariates. The same priors can be placed on the continuous covariate means $\boldsymbol{\mu}_k$ but we do not find this necessary in the present work. 

\begin{enumerate}

\item \textbf{Laplace ($\ell_1$) Prior}: 
We assume that the means $\pi_{jk}$ follow zero-mean Laplace distributions and are independent across mixture components and covariates. The negative log-likelihood is therefore proportional to the $\ell_1$ norm 
\begin{equation}\label{eqn:Laplace}
\Omega(\boldsymbol{\pi}) = \sum_{j\in\text{disc}} \sum_{k=1}^K \lvert \pi_{jk} \rvert,
\end{equation}
where the summation over $j$ is restricted to the discrete covariates.

\item \textbf{Group $\ell_{1,2}$ Prior}: It may further be the case that some covariates are non-informative of group membership, in which case the means $\pi_{jk}$ 
should be zero across all groups $k$ and follow the group $\ell_{1,2}$ distribution \citep{MarlinSM2009},  
similar to the group lasso \citep{YuanL2006}. The corresponding negative log-likelihood is 
\begin{equation}\label{eqn:groupSparse}
\Omega(\boldsymbol{\pi}) = \sum_{j\in\text{disc}} \sqrt{\sum_{k=1}^K \lvert \pi_{jk} \rvert^2}.
\end{equation}
\end{enumerate}

\subsection{Treatment Outcome Model}
\label{sec:outcome}

We model the enhanced or diminished treatment effect in a subgroup through the following relationships.  In the case where $Y$ is binary, its mean is equal to the probability of $Y=1$.  We define the latter using the logistic sigmoid function $g$ to be 
\begin{equation}\label{eqn:outcome}
  p(Y=1 \given \mathbf{x}, t, Z=k; \mathbf{w}_t, \gamma_k) = g\bigl( f(\mathbf{x}; \mathbf{w}_t) + \gamma_k t\bigr), 
\end{equation}
where $f(\mathbf{x}; \mathbf{w}_t)$ is a function of $\mathbf{x}$ parametrized by $\mathbf{w}_t$, $t=0,1$.  The term $\gamma_k t$ represents the main effect due to treatment and the coefficient $\gamma_k$, i.e.~the size of the effect, depends on the group membership $Z=k$. 
The parameters $\mathbf{w}_t$ are allowed to be different for $t=0$ and $t=1$ to better account for differing covariate distributions $p(\mathbf{x} \given t)$ between the two treatment groups, a.k.a.~selection bias.  In the case of continuous $Y$, we replace $g$ with the identity function as follows:
\begin{equation}\label{eqn:outcomeCont}
  \mathbb{E}[Y=1 \given \mathbf{x}, t, Z=k; \mathbf{w}_t, \gamma_k] = f(\mathbf{x}; \mathbf{w}_t) + \gamma_k t. 
\end{equation}

The simplest choice for function $f(\cdot)$ is linear, i.e., two linear functions $\mathbf{w}_0^\top \mathbf{x}$ and $\mathbf{w}_1^\top \mathbf{x}$.  
In practice, however, the outcome may have a highly nonlinear dependence on the covariates. To accommodate nonlinear covariate interactions and thus better adjust for confounding, we also allow $f$ to be a nonlinear function. In this paper, we experiment with one- and two-hidden-layer feedforward neural networks with ReLU activations.  Outcomes under $t=0$ and $t=1$ are produced by two different heads of the network, following \cite{shalit2016estimating, louizos2017causal, johansson2016learning}.  
Even in the nonlinear case, the assignment of an individual to a subgroup is still described by a mixture model and directly interpretable in terms of the original feature representation, thus preserving interpretability of the discovered subgroups. 

It is possible to regularize the outcome models \eqref{eqn:outcome}, \eqref{eqn:outcomeCont} with $\ell_2$ or $\ell_1$ regularization $\Lambda(\mathbf{w}_t)$, which is equivalent to adding a normal or Laplace prior on the parameter $\mathbf{w}_t$.  In this work however, we use weight decay instead as discussed in Section~\ref{sec:inference:ELBO}.

\section{Inference}
\label{sec:inference}

We would like to fit our proposed model to a given observational dataset $\mathcal{D}$.  Denote by $\mathbf{\Theta} = (\{\boldsymbol{\mu}_k, \Sigma_k, \boldsymbol{\pi}_k, \gamma_k\}_{k=1}^K, \mathbf{w})$ the set of all parameters of the model. 

We have considered two approaches: maximizing the joint likelihood $p(\mathbf{x}_i, t_i, y_i; \mathbf{\Theta})$, and maximizing the conditional likelihood $p(y_i | \mathbf{x}_i, t_i; \mathbf{\Theta})$.  The joint and conditional likelihoods can be related as follows:  
%
\begin{equation}\label{eqn:LLjoint}
  \sum_{i=1}^N \ln p(\mathbf{x}_i, t_i, y_i)
    = \sum_{i=1}^N \left[ \ln p(\mathbf{x}_i) + \ln p(t_i \given \mathbf{x}_i) + \ln p(y_i\given \mathbf{x}_i, t_i) \right].
\end{equation}
The conditional likelihood can be further expanded as 
\begin{equation}\label{eqn:LLcond}
\ln p(y_i\given \mathbf{x}_i, t_i) = \ln \left(\sum_{k=1}^K p(z_i=k \given \mathbf{x}_i) p(y_i\given \mathbf{x}_i, t_i, z_i=k) \right),
\end{equation}
where we have used the conditional independence of $Z$ and $T$ given $\mathbf{X}$ in the first factor on the right-hand side.  The resulting first factor $p(z_i=k \given \mathbf{x}_i)$ as well as the term $p(\mathbf{x}_i)$ in \eqref{eqn:LLjoint} depend only on the mixture model \eqref{eqn:Z}--\eqref{eqn:Xdisc}, to wit $p(\mathbf{x}_i) = \sum_{k=1}^K p(z_i=k) p(\mathbf{x}_i \given z_i=k)$ and $p(z_i=k \given \mathbf{x}_i) = p(z_i=k) p(\mathbf{x}_i \given z_i=k) / p(\mathbf{x}_i)$.  The second factor on the right-hand side of \eqref{eqn:LLcond} depends only on the outcome model \eqref{eqn:outcome}, \eqref{eqn:outcomeCont}.  The remaining term $p(t_i \given \mathbf{x}_i)$ in \eqref{eqn:LLjoint} depends on the propensity model.  Since this is the only place where the propensity model appears, 
its inference is separable from the remainder of the problem, as claimed in Section~\ref{sec:genModel}.  We do not discuss propensity modeling further as it is not the focus of this work.

    
Although maximizing the joint likelihood \eqref{eqn:LLjoint} results in some closed-form expressions 
and accordingly easier inference of parameters, we have observed in practice that maximizing the conditional likelihood \eqref{eqn:LLcond} 
has superior performance in estimating the potential outcomes $Y(t)$ and treatment effects. 
Therefore, we pursue this discriminative approach in this work.  We do however include the sparsity-inducing prior on the parameters $\boldsymbol{\pi}_k$ 
discussed in Section~\ref{sec:sparsity}. 
The full objective function is therefore 
\begin{equation}\label{eqn:objective}
    \sum_{i=1}^N \ln p(y_i\given \mathbf{x}_i, t_i; \mathbf{\Theta}) - \lambda \Omega(\mathbf{\boldsymbol{\pi}}), 
\end{equation}
where $\lambda$ 
controls the strength of the prior.

\subsection{Evidence Lower Bound (ELBO) Optimization}
\label{sec:inference:ELBO}

Instead of optimizing the conditional log-likelihood in \eqref{eqn:objective} directly using a gradient method, we choose to lower bound the likelihood with a variational approximation, more commonly known as the Evidence Lower Bound (ELBO) \cite{blei2017variational}. 
For any variational distribution $q(Z)$ over the latent variable $Z$, we have 
\begin{align}
   \ln p(y_i\given \mathbf{x}_i, t_i; \mathbf{\Theta})
    &= \ln \sum_{k=1}^{K}    p(y_i, z_i=k\given \mathbf{x}_i, t_i; \mathbf{\Theta})\nonumber \\
    &=  \ln \left( \mathbb{E}_{q} \Big[ \frac{p(y_i, z_i\given \mathbf{x}_i, t_i; \mathbf{\Theta})}{q(Z)} \Big] \right)\nonumber\\
    &\geq  \mathbb{E}_{q} \Big[ \ln \frac{p(y_i, z_i\given \mathbf{x}_i, t_i; \mathbf{\Theta})}{q(Z)} \Big]    \label{eq:ELBOeq}
\end{align}
using Jensen's inequality.  Now, replacing 
$q(Z)$ with $p(z_i|\mathbf{x}_i;\mathbf{\Theta})$ and using \eqref{eqn:LLcond} (and $Z \independent T \given \mathbf{X}$ from Figure~\ref{fig:plate}), we obtain
\begin{align}
\text{ELBO}(y_i,\mathbf{x}_i, t_i; \mathbf{\Theta}) 
&= \sum_{k=1}^K  p(z_i=k \given \mathbf{x}_i;\mathbf{\Theta}) \ln{p(y_i \given \mathbf{x}_i, t_i, z_i=k; \mathbf{\Theta})}. \label{eqn:ELBO}
\end{align}

We hence substitute \eqref{eqn:ELBO} in place of $\ln p(y_i\given \mathbf{x}_i, t_i; \mathbf{\Theta})$ in \eqref{eqn:objective} and proceed to maximize the objective function 
using the Adam gradient method \cite{kingma2014adam}, a variant of stochastic gradient descent that is a popular choice for non-convex functions like neural networks.  The same method is used for both linear and nonlinear $f$ in \eqref{eqn:outcome}, \eqref{eqn:outcomeCont}.  As noted above, the first factor $p(z_i=k \given \mathbf{x}_i;\mathbf{\Theta})$ in \eqref{eqn:ELBO} depends only on the mixture model parameters in \eqref{eqn:Z}--\eqref{eqn:Xdisc} while the second factor depends only on the outcome model parameters in \eqref{eqn:outcome}, \eqref{eqn:outcomeCont}.  We enable ``weight decay'' \cite{krogh1992simple} on the parameters $\mathbf{w}_t$ as a form of regularization.  For tractability, we compute the ELBO only over a fixed-size mini-batch of the data before each parameter update.  Additional details on the algorithm and parameter initialization can be found in Appendices~\ref{sec:inference:init} and \ref{sec:expt:fit} in the supplement.

We also considered an expectation-maximization (EM) algorithm to maximize \eqref{eqn:objective} as an alternative to ELBO.  Our experience however was that ELBO provided better fit in terms of Log-Likelihood and heterogeneous effect estimates in terms of the metric reported in Section~\ref{sec:expt:hetero}. A full description of the EM method and a comparison to the ELBO optimization is deferred to the Appendix.

\begin{figure*}[!htbp]

  \begin{subfigure}[t]{0.4\textwidth}
  \centering
  \includegraphics[width=\linewidth]{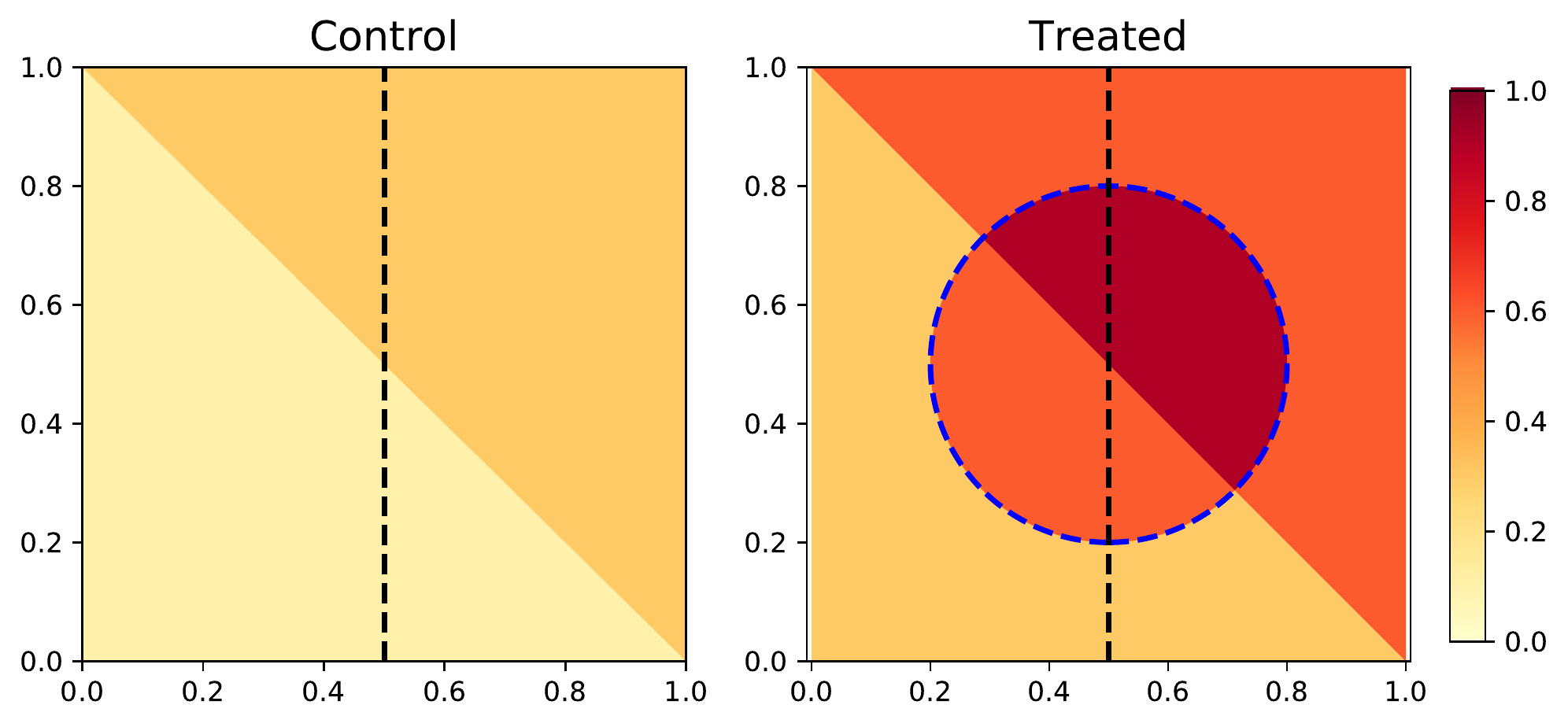}
  \caption{Potential outcome probabilities $p(Y(0)=1 \given \mathbf{X})$ (control) and $p(Y(1)=1 \given \mathbf{X})$ (treated). \small Treatment propensity is greater to the right of the black dashed line, while the blue dashed line denotes the region with enhanced effect.}
    \label{fig:syndata:outcome}
    \end{subfigure}
     \begin{subfigure}[t]{0.19\textwidth}
      \centering
  \includegraphics[width=\linewidth]{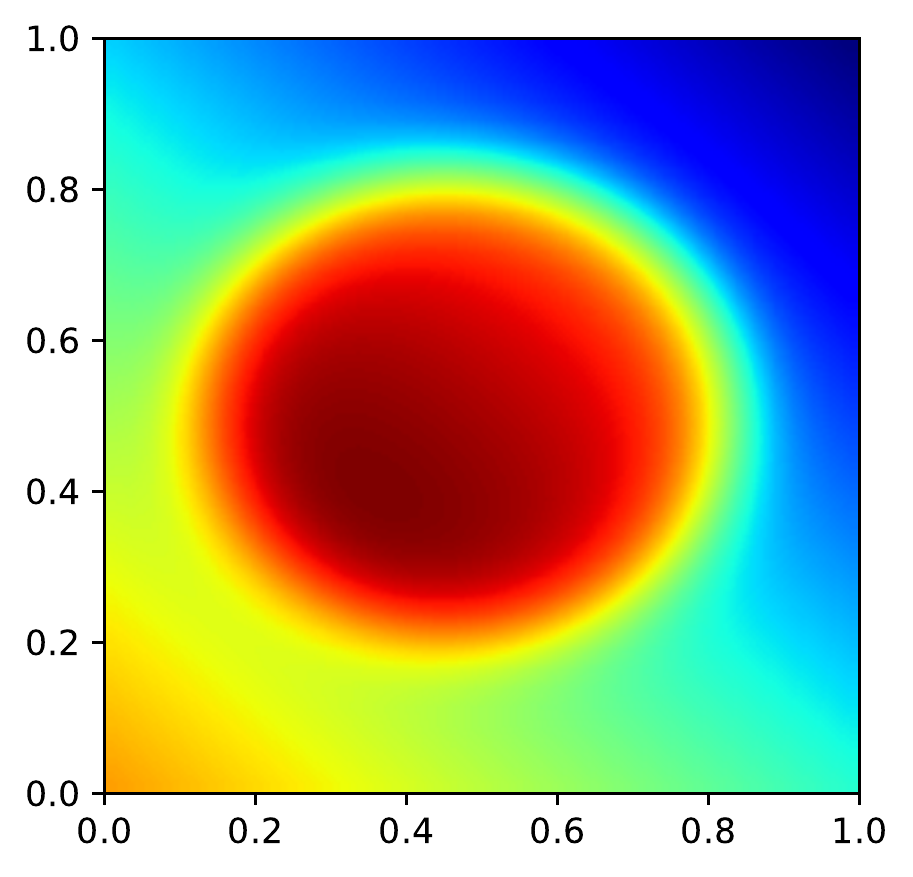}
  \caption{\method{}-Lin subgroup}
    \label{fig:syndata:HTMM}
    \end{subfigure}
    \begin{subfigure}[t]{0.19\textwidth}
      \centering
  \includegraphics[width=\linewidth]{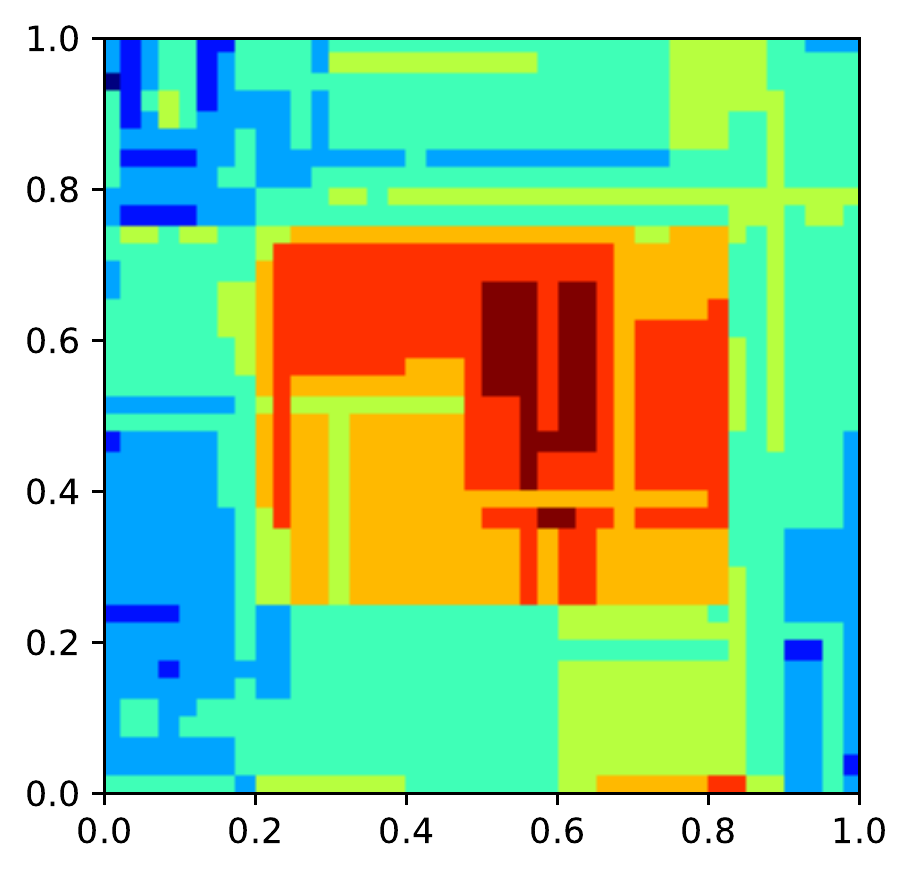}
    \caption{VT-C subgroup}
  \label{fig:syndata:VT-C}
    \end{subfigure}
    \begin{subfigure}[t]{0.19\textwidth}
      \centering
  \includegraphics[width=\linewidth]{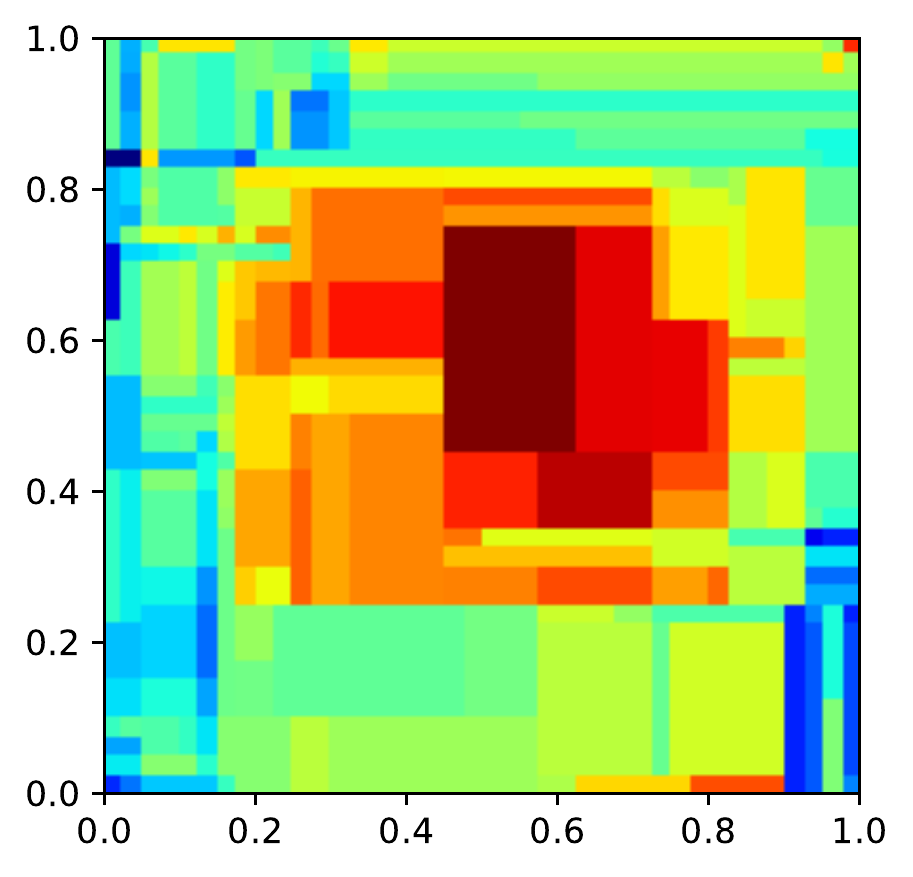}
  \caption{VT-R subgroup}
  \label{fig:syndata:VT-R}    
    \end{subfigure}
    \label{fig:syndata}
  \caption{\dssyn{} dataset and enhanced effect subgroups discovered by \method{} and Virtual Twins (VT).}
\end{figure*}

\section{Dataset Descriptions}
\label{sec:data}

We demonstrate the performance of \method{} on a synthetic dataset, the semi-synthetic Infant Health and Development Program (\dsihdp{}) dataset, and a real-world dataset on opioids. These datasets are described further below.\\

\noindent \dssyn{}:  We take $
\mathbf{X} = (X_0, X_1) \in \mathbb{R}^2$ and sample it from a uniform distribution over $\mathcal{X} = [0,1]^2$. In order to simulate the selection bias inherent in observational studies, the treatment variable depends on $\mathbf{X}$ as $T \sim \text{Bernoulli}(0.4)$ for $x_0 < 0.5$ and $\text{Bernoulli}(0.6)$ for $x_0 > 0.5$.  The potential outcomes $Y(0)$ and $Y(1)$ are also Bernoulli with means given by the functions of $\mathbf{X}$ 
shown in Figure~\ref{fig:syndata:outcome}.  The figure shows that $p(Y(1)=1 \given \mathbf{X}) > p(Y(0)=1 \given \mathbf{X})$, i.e.~treatment increases the 
probability of positive outcome.  Note that under the conditional exchangeability assumption we have $p(Y(t)=1 \given \mathbf{X}) = p(Y=1 \given T=t, \mathbf{X})$.  We model the effect of the confounders $\mathbf{X}$ by assigning higher probability to the upper triangular region of $\mathcal{X}$. This together with the distribution of $T$ imply that individuals who are more likely to have positive outcome regardless of treatment (upper triangle) are also more likely to receive treatment (right half-square). Lastly, we model the enhanced treatment effect group as a circular region $\mathcal{S} = \{x: \lVert x-c \rVert_{2} < r \}$, where $p(Y(1)=1 \given \mathcal{S}) > p(Y(1)=1 \given \mathcal{X}\backslash\mathcal{S})$. We set $c = ( \frac{1}{2}, \frac{1}{2})$ and $r = \frac{1}{4}$.  A total of $1000$ samples $(\mathbf{x}_i, t_i, y_i)$ were generated as described above.
\\

\noindent \textbf{\textsc{IHDP (Semi-Synthetic)}}: The IHDP dataset has gained popularity in the causal inference literature dealing with heterogenous treatment effects \cite{alaa2017bayesian,shalit2016estimating,louizos2017causal,hill2011bayesian}. The original data includes 25 real covariates and comes from a randomized experiment to evaluate the benefit of IHDP on IQ scores of three-year-old children. A selection bias was introduced by removing some of the treated population, thus resulting in 608 control patients and 139 treated (747 total). The outcomes were simulated using the standard non-linear `Response Surface B' as described in \cite{hill2011bayesian}. \\

\begin{table}
\centering
\begin{tabular}{l|l|c}
\hline
Total Covariate Dimension & \textbf{1226}&3 Continuous,\\ 
 & &1223 Binary\\ \hline
ICD-9 Diagnostic Codes & 1013&Binary\\ \hline
CPT Procedure Codes & 171 &Binary\\ \hline
Hand-Crafted Comorbidities & 41&Binary\\ \hline
Daily Morphine Equivalent, &3 &Continuous\\ 
Total Number of Visits, Age & & \\ \hline

\end{tabular}\\
\vspace{1em}
\begin{tabular}{l|l|l|l}
&Addicted &Not-Addicted&Total\\
&($Y$=1) &($Y$=0)&\\ 
\hline
Treated ($T$=1)&2060&19983&\textbf{22043}\\
Control ($T$=0)&7269&176156&\textbf{183425}\\ \hline
Total&\textbf{9329}&\textbf{196139}&\textbf{205468}\\ 
\hline
\end{tabular}
\caption{\dsopioid{} Dataset Statistics}
\label{tbl:statistics}
\end{table}

\noindent \textbf{\textsc{Opioid}}: We sampled a sub-population consisting of healthcare claims for five million patients from the MarketScan Commercial claims database. These claims describe patients' medical histories, including both inpatient admissions and outpatient services.  Diagnoses, procedures, prescriptions and dosages are recorded.  We follow the cohort selection procedure outlined in \citet{zhangexploring} to filter patients based on several criteria. 

For each patient in our final cohort, we create a feature vector that includes basic demographic information such as age, gender and geographic region. We also included a predefined set of procedures along with diagnostic codes which are associated with opioids and/or addiction, based on input from a physician.


We label all patients who received addiction diagnoses 
and patients who continued use of opioids for more than one year after the initial prescription 
as belonging to the positive (adverse outcome) class. Patients who discontinued opioid use within one year of initial treatment were labeled as negative. We use the terms ``addicted'' and ``not addicted'' as shorthand for these outcomes. Patients prescribed natural or semi-synthetic opioids are considered the control group, whereas patients administered synthetic opioids are considered the treated group. Table \ref{tbl:statistics} summarizes the basic statistics of this dataset.

\section{Experimental Results}
\label{sec:expt}

We evaluated the proposed \method{} quantitatively on two tasks, prediction of heterogeneous treatment effects and identification of subgroups with enhanced or reduced effect.  These results are discussed in Sections~\ref{sec:expt:hetero} and \ref{sec:expt:subgroup} respectively in comparison to existing methods, focusing on those that also estimate heterogeneous effects in an interpretable manner.  Methods used in the comparison are described in Section~\ref{sec:expt:comp} and parameter selection details are in Appendix~\ref{sec:expt:fit}.  In Section~\ref{sec:expt:interpret}, we provide qualitative results for the \dsopioid{} dataset on the features the model discovers as characteristics of ``at-risk'' individuals, i.e.~those in enhanced effect subgroups.

\subsection{Algorithms in Comparison}
\label{sec:expt:comp}

We have considered Virtual Twins (VT) \cite{foster2011subgroup}, QUINT \cite{Dusseldorp2014}, and SIDES \cite{lipkovich2011subgroup} among methods that identify subgroups with different treatment effects.  We implemented two versions of VT in which the treatment effect is modeled by a decision tree classifier (VT-C) or regressor (VT-R). For VT-C, to better represent the continuous-valued treatment effect (which is a difference in probabilities even if $Y$ is binary), we use a collection of decision tree classifiers obtained by applying different thresholds to the treatment effect. 

For QUINT and SIDES, we utilized the standard R implementations and performed extensive hyperparameter tuning. However both QUINT and SIDES failed to recover any subgroups on \dssyn{} and \dsopioid{} and we thus did not consider them further.  For QUINT, the likely reason is that its assumption of a subgroup with diminished effect is not always met, whereas for SIDES, there may be a numerical issue in how it discretizes continuous covariates.

In terms of methods that only predict heterogeneous effects and do not identify subgroups, we also compare our method with some common approaches in Table \ref{tbl:PEHE}. Here Linear-1 corresponds to a single ordinary least squares (for continuous outcomes) or logistic regression (for binary outcomes) for both factual and counterfactual outcomes. In the case of Linear-2, we fit two separate linear models to the control and treated populations to better accommodate selection bias and confounding. The other baselines, $k$-NN, GP, and CFRF are non-parametric versions of this approach where the estimators of the factual and counterfactual outcomes are $k$-nearest neighbours, Gaussian processes with a linear kernel and Random Forests respectively. 



\subsection{Heterogeneous Effect Estimation}
\label{sec:expt:hetero}

We first evaluate our performance on estimation of the CATE ($\mathbb{E}[Y(1) - Y(0) \given \mathbf{X}]$). A popular metric for this evaluation is the {\it Precision in Estimating Heterogenous Effects} (PEHE). The PEHE is defined as
\begin{equation*}
\text{PEHE} = \frac{1}{n} \sum^{n}_{i=1}  \left(f_1(\mathbf{x_i}) - f_0(\mathbf{x}_i) - \mathbb{E}[Y(1) - Y(0) \given \mathbf{X} = \mathbf{x}_i] \right)^2.
\end{equation*}
Here $f_1(\cdot)$ and $f_0(\cdot)$ are the estimated potential outcomes under treatment and control, respectively.

Table \ref{tbl:PEHE} compares the performance of \method{} against the methods described in Section \ref{sec:expt:comp} on both in-sample PEHE (corresponding to a retrospective study) computed on the training data, and out-of-sample PEHE computed on held-out test data. \method-MLP and \method-Lin refer to the proposed approach with $f$ in \eqref{eqn:outcome}, \eqref{eqn:outcomeCont} as a multilayer perceptron and linear function respectively to model the effect of confounders on the outcome.

HEMM consistently outperforms these standard causal inference baselines. GP and Linear-2 perform close to HEMM on \dssyn{}. We noticed that when a larger sample of data points is available to VT-R, its performance increases dramatically. However, its performance drops 
in higher-dimensional settings as in the case of \dsihdp{} and \dsopioid{}; this is expected with methods involving non-parametric regression.

\begin{figure*}[!t]
  \centering
  \begin{subfigure}{0.33\textwidth}
  \includegraphics[width=1\linewidth]{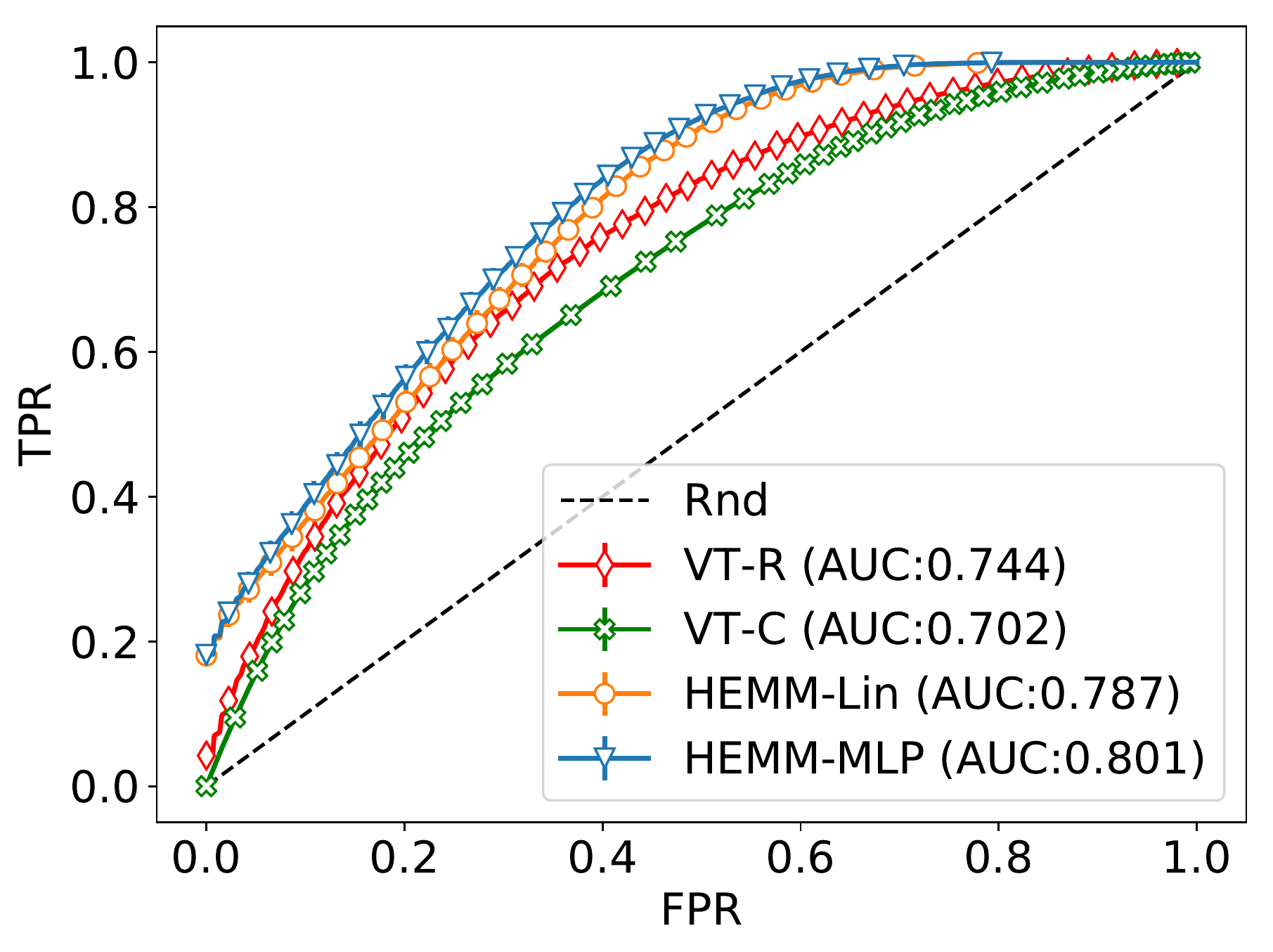}
  \caption{\dssyn{}}
  \label{fig:subgroupdiscovery:syn}
  \end{subfigure}
  \begin{subfigure}{0.33\textwidth}
  \includegraphics[width=1\linewidth]{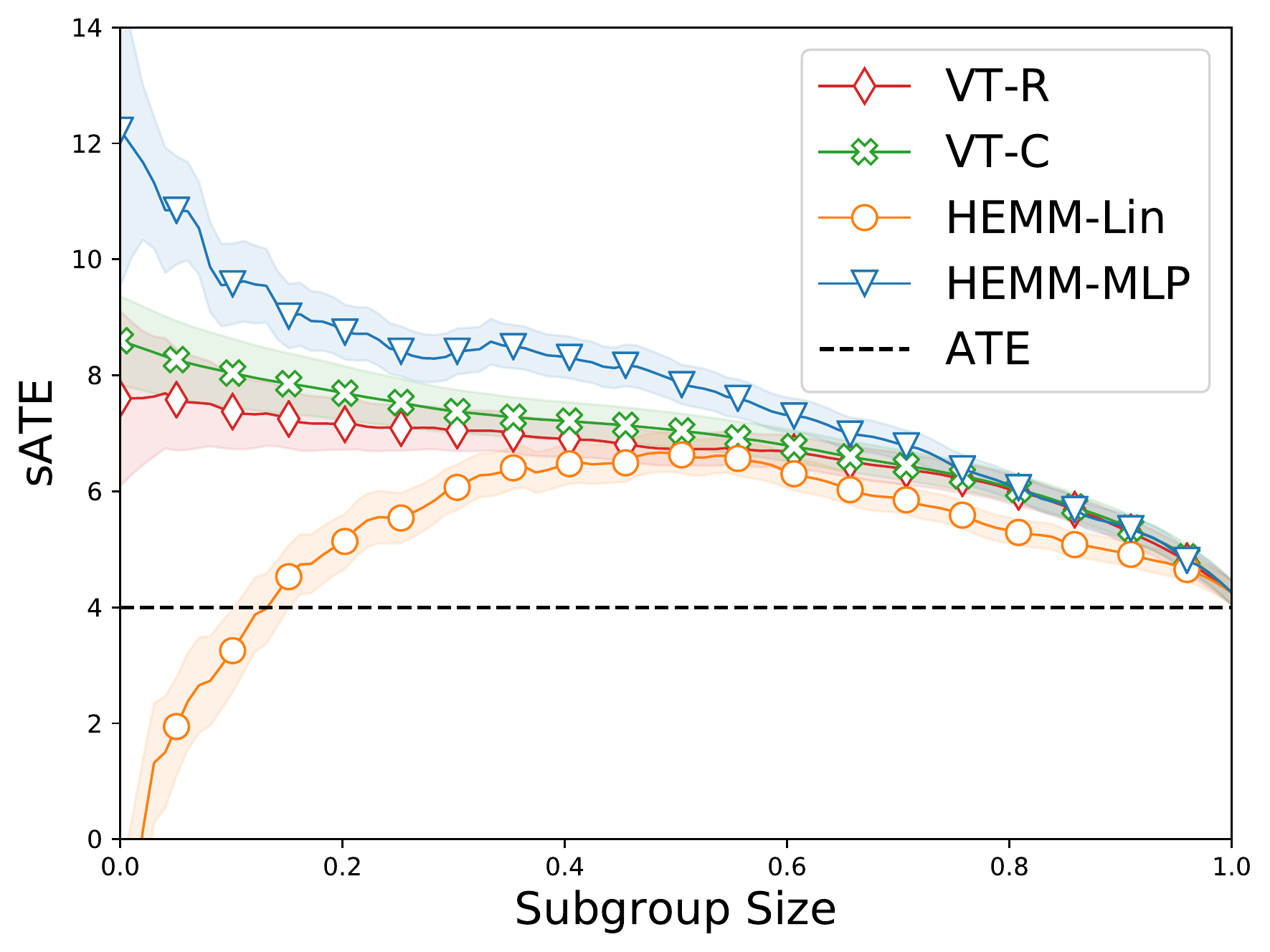}
  \caption{\dsihdp{}}
  \label{fig:subgroupdiscovery:ihdp}
  \end{subfigure}
  \begin{subfigure}{0.33\textwidth}
  \includegraphics[width=1\linewidth]{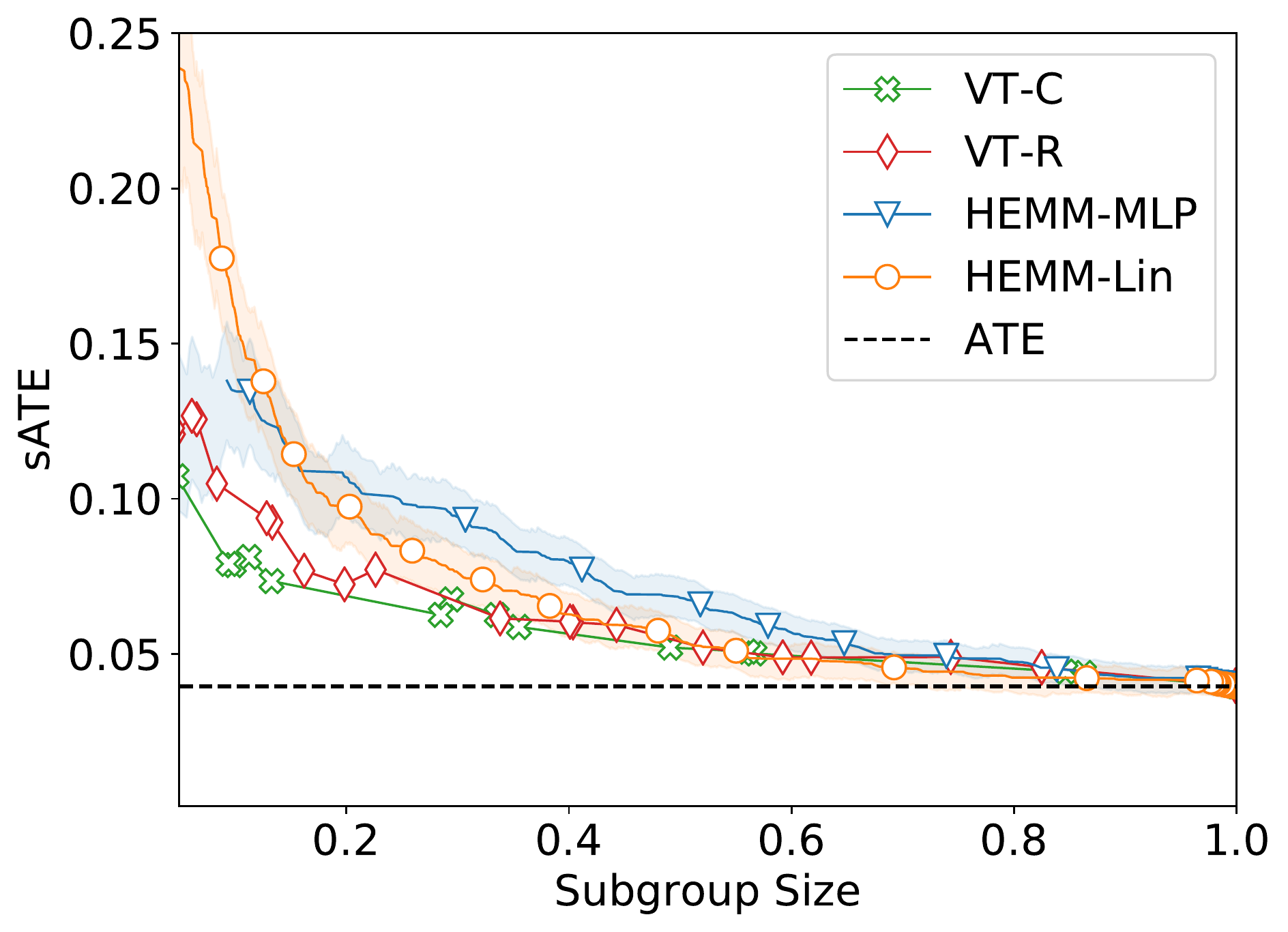}
  \caption{\dsopioid{}}
  \label{fig:subgroupdiscovery:opioid}
  \end{subfigure}
  \caption{Performance of the proposed \method{} and Virtual Twins on the subgroup discovery task. For
  \dssyn{} data, we have access to ground truth labels for the enhanced treatment effect group and hence compare performance using the ROC. For the \dsihdp{} and \dsopioid{} datasets, we compare average treatment effect (ATE) estimates within the identified subgroup as a function of subgroup size (as a fraction of the population).}
    \label{fig:subgroupdiscovery}
\end{figure*}





 
\definecolor{Gray}{gray}{0.8}
\begin{table}[!htbp]
\centering
 \resizebox{.5\textwidth}{!}{\begin{tabular}{lcc|cc}
\toprule
&\multicolumn{2}{c}{\dssyn{}}&\multicolumn{2}{c}{\dsihdp{}}\\ \midrule
&In-sample&Out-Sample&In-sample&Out-Sample\\ \hline \rowcolor{Gray}
\textbf{HEMM-MLP}&$\mathbf{0.101\pm10^{-3}}$&$\mathbf{0.102\pm10^{-3}}$&$\mathbf{1.6\pm0.10}$&$\mathbf{1.8\pm0.10}$\\ \rowcolor{Gray}
\textbf{HEMM-Lin}&$\mathbf{0.116\pm10^{-3}}$&$\mathbf{0.116\pm10^{-3}}$&$\mathbf{2.8\pm0.32}$&$\mathbf{2.9\pm0.33}$\\ \hline
Linear-1&$0.278\pm10^{-3}$&$0.278\pm10^{-3}$&$7.9\pm0.46$&$7.9\pm0.47$\\
Linear-2&$0.106\pm10^{-3}$&$0.107\pm10^{-3}$&$2.3\pm0.18$&$2.4\pm0.21$\\ \hline
$k$-NN&$0.210\pm10^{-3}$&$0.210\pm10^{-3}$&$3.2\pm0.12$&$4.2\pm0.22$\\ 
GP&$0.106\pm10^{-3}$&$0.107\pm10^{-3}$&$2.1\pm0.11$&$2.3\pm0.14$\\ \hline
CFRF&$0.146\pm10^{-3}$&$0.142\pm10^{-3}$&$2.7\pm0.31$&$3.3\pm0.72$\\
VT-R&$0.130\pm10^{-3}$&$0.130\pm10^{-3}$&$2.5\pm0.26$&$2.9\pm0.51$\\ \bottomrule
\end{tabular}}
\caption{$\sqrt{\text{PEHE}}$ values in estimating heterogeneous effects. Error represents 95\% confidence interval of multiple Monte Carlo initializations.}
\label{tbl:PEHE}
\end{table}

\subsection{Subgroup Identification}
\label{sec:expt:subgroup}

\dssyn{}:  In the case of synthetic data, the subgroup with enhanced treatment effect is known.  We first visualize the performance of \method{} and VT in identifying this subgroup. 
For \method{}-Lin, Figure~\ref{fig:syndata:HTMM} shows the estimated probability $p(Z\given\mathbf{X})$ of belonging to the enhanced effect subgroup evaluated on the test set. The true circular region is recovered well.
Figures~\ref{fig:syndata:VT-C} and \ref{fig:syndata:VT-R} plot the VT-C and VT-R predictions of CATE on the test set.  For VT-C, the prediction represents an average over the collection of decision tree classifiers, while for VT-R, it is simply the output of the decision tree regressor. While the difficulty in reproducing the circular shape is expected for decision trees, the enhanced effect estimates are also less uniform than in Figure~\ref{fig:syndata:HTMM}. 

We also evaluate subgroup identification more quantitatively by treating it as a problem of classifying whether or not points in the test set belong to the enhanced effect subgroup.  ROC curves may then be plotted as in Figure~\ref{fig:subgroupdiscovery:syn}. For \method{}, the ROC is traced by varying the threshold on the probability $p(Z\given\mathbf{X})$ of being in the enhanced effect subgroup (shown in Figure~\ref{fig:syndata:HTMM}).  Similarly for VT-C and VT-R, the threshold on the CATE estimates (Figures~\ref{fig:syndata:VT-C} and \ref{fig:syndata:VT-R}) is varied.  \method{} has higher ROCs than VT on this example, in line with Figures~\ref{fig:syndata:HTMM}--\ref{fig:syndata:VT-R}.  There is little difference between \method{}-Lin and \method{}-MLP since the dependence on the covariates $\mathbf{X}$ in Figure~\ref{fig:syndata:outcome} is simple and complex adjustment is not needed.

\noindent\dsihdp{} and \dsopioid{}:  For the other two datasets, we conduct a relative comparison with VT since we lack data on ground truth subgroups. 
The evaluation involves two steps. First, we assign individuals to an enhanced effect subgroup of varying size. (The same procedure can be used for a diminished effect subgroup but we omit the results due to space.)  For \method{}, we choose the subgroup $k$ with the largest main effect $\gamma_k$ 
and vary the threshold applied to the corresponding membership probability $p(Z=k\given \mathbf{X})$ returned by the model. 
For VT-C and VT-R, we vary the threshold applied to the CATE estimates, either the composite estimate of the decision tree classifiers or the regressor estimate, the same quantities as for the synthetic data. 


In the second step, we build a propensity score model (an estimator of treatment propensity $p(T=1|\mathbf{X})$) to estimate the average treatment effect (ATE) conditioned on belonging to the enhanced treatment effect subgroup defined in the first step. For the propensity score model $e(\mathbf{X})$, we fit a random forest, for which parameter tuning is performed on the \textsc{dev} set. We then use the inverse probability of treatment weighting (IPTW) estimator \citep{imbens2004nonparametric} of the ATE within a subgroup $\mathcal{S}$ as follows:%
\begin{equation}
\hat{\tau}_{\mathcal{S}} =\frac{1}{|\mathcal{S}|}\sum_{i \in \mathcal{S}} \left( \frac{y_i t_i}{e(\mathbf{X}_i)} - \frac{y_i (1-t_i)}{1-e(\mathbf{X}_i)} \right).
\label{eqn:subgroupiptwestimator}
\end{equation}%
IPTW estimation is used for both \method{}- and VT-defined subgroups to be consistent.

Figures~\ref{fig:subgroupdiscovery:ihdp} and \ref{fig:subgroupdiscovery:opioid} plot subgroup ATE versus subgroup size (as a fraction of the population) as the threshold for subgroup assignment is varied. When the subgroup is the entire population at size $1.0$, all curves meet at the population ATE (dashed line).  Since we have selected the enhanced effect subgroup, the curves are then expected to increase as the subgroup is restricted to individuals with larger treatment effects. The fact that this increase is nearly monotonic for \method{}-MLP 
is evidence for the validity of the discovered subgroup, since the IPTW estimator used here is an independent check on the treatment effect model \eqref{eqn:outcome}, \eqref{eqn:outcomeCont} used by \method{}. Compared to VT, the subgroups identified by \method{}-MLP have higher ATE. 
This suggests that for a given subgroup size, \method{}-MLP is better at grouping together individuals with more enhanced effects. 
\method{}-Lin on the other hand displays contrasting performances.  On IHDP in Figure~\ref{fig:subgroupdiscovery:ihdp}, the estimated ATE actually decreases for subgroup sizes less than $0.5$, likely due to the inadequacy of a linear model to adjust for confounding and accurately estimate CATE.  In Figure~\ref{fig:subgroupdiscovery:opioid} however, the ATE does increase monotonically and faster than for VT.


\subsection{Interpretation of the \dsopioid{} Enhanced Effect Subgroup}
\label{sec:expt:interpret}

\begin{table}[!t]

\centering
\small
 \resizebox{.5\textwidth}{!}{
\begin{tabular}{ | p{2in} | p{2in} | }
\hline \rowcolor{Gray}
  \multicolumn{1}{|c|}{\textbf{Musculoskeletal System}} & \multicolumn{1}{|c|}{\textbf{Nervous System}} \\ \hline
1.0 spinal curve \small{(kyphosis, lordosis, scoliosis)} & 1.0 extrapyramidal diseases/movmt. disorders \\ 
1.0 ankle fracture & 1.0 idiopathic peripheral neuropathies \\ 
  1.0 sprains/strains of hand and wrist & 1.0 headaches \\ \hline \rowcolor{Gray}
  \multicolumn{1}{|c|}{\textbf{Integumentary System}} & \multicolumn{1}{|c|}{\textbf{Endocrine System}} \\ \hline
  1.0 cellulitis and abscess of finger and toe & .70 simple and unspecified goiter \\ 
  1.0 local skin infections & .67 other endocrine disorders \\ 
  1.0 psoriasis and similar disorders & .65 thyrotoxicosis with or without goiter \\ \hline \rowcolor{Gray}
  \multicolumn{1}{|c|}{\textbf{Reproductive System}} & \multicolumn{1}{|c|}{\textbf{Digestive and Excretory Systems}} \\ \hline
  1.0 female infertility & .71 benign neoplasm of intestinal tract \\ 
  .82 testicular dysfunction & .69 inguinal hernia \\ 
  .82 disorders of penis & .58 diverticulitis \\ \hline \rowcolor{Gray}
  \multicolumn{1}{|c|}{\textbf{Circulatory System} }& \multicolumn{1}{|c|}{\textbf{Immune System}} \\ \hline
  1.0 hypertensive heart disease & .56 immunization \\ 
  .79 other disorders of circulatory system & .54 strep throat and scarlet fever \\ 
  .70 cardiac dysrhythmias & .52 bacterial infections in other conditions \\ \hline \rowcolor{Gray} 
  \multicolumn{1}{|c|}{\textbf{Nutrition}} & \multicolumn{1}{|c|}{\textbf{Visual System}} \\ \hline
  1.0 BMI & .87 keratitis \\ 
  1.0 b-complex deficiency & .60 other disorders of eye (epi)scleritis \\ 
  .71 disorder of electrolyte/acid-base balance & .57 visual disturbances \\ \hline \rowcolor{Gray}
  \multicolumn{1}{|c|}{\textbf{Auditory System}} & \multicolumn{1}{|c|}{\textbf{Psychology}} \\ \hline 
  .53 vertiginous syndrome/vestibular disorder & 1.0 suspected mental health condition \\ 
  .50 otitis media/eustachian tube disorders & .58 adjustment reaction \\ 
  .44 disorders of pinna and mastoid process & .55 nondependent abuse of drugs \\ \hline \rowcolor{Gray}
  \multicolumn{1}{|c|}{\textbf{Digestive System (upper/oral)}} & \multicolumn{1}{|c|}{\textbf{Respiratory System}} \\ \hline
  .77 hernia, abdominal cavity w/o obstruction & 1.0 other diseases of respiratory tract  \\ 
  .77 dentofacial anomalies of jaw & .74 deviated nasal septum \\ 
  .76 diseases of oral soft tissues & .69 influenza \\ \hline
\end{tabular}
}
\caption{Top features of the enhanced effect subgroup $k$ discovered by \method{}-MLP on the \dsopioid{} dataset. The numbers are the ratios $\pi_{jk} / \sum_{k'} \pi_{jk'}$, where $1/2$ represents no increase in prevalence over the other subgroup ($K=2$).}
\label{tab:nonlinearinterpretation}

\end{table}

We now turn our focus to the motivating application of opioids and analyze key characteristics of the enhanced effect subgroup, i.e.~those patients at greater risk of adverse outcomes when treated initially with synthetic opioids.  To interpret these features, we collaborated with a subject matter expert (SME) with a PhD in cognitive neuroscience and a clinical research emphasis in chronic pain conditions and treatments, including opioids. 
Table \ref{tab:nonlinearinterpretation} 
shows the top features of the enhanced effect subgroup as identified by \method{}-MLP. 
The features are organized by general bodily system and sorted in descending order of prevalence relative to the other subgroup (see table caption); the selection of 3 features in each system was arbitrary and chosen primarily for simplicity and space constraints.  

Patients with a history of chronic conditions in general, as well as chronic pain conditions more specifically, are at an increased risk for addiction.  Many of the chronic conditions in Table 3, e.g.~heart disease (circulatory system), psoriasis (integumentary system), and BMI/obesity (nutrition) also appear in the CDC opioid prescribing guidelines \cite{dowell2016CDC} or have extensive literature linking them to increased risk for long-term pain, either intrinsic to the condition or due to needed medical procedures that are more likely to expose patients to opioids \cite{glanz2018prediction}. 
For example, numerous papers show a link between increased body-mass index (BMI) and increased pain intensity and duration (with anti-correlations between BMI and pain recovery) \cite{okifuji2015association}, and obesity has also been associated with higher initial opioid doses 
\cite{kobus2012correlates}.  Additionally, 
the chronic nutritional deficiencies and imbalances shown in Table~\ref{tab:nonlinearinterpretation} have been linked to acute but intense muscle spasms as well as peripheral polyneuropathies and paresthesias 
(see e.g.~\cite{mostacci2018nutraceutical}), pain disorders which also show up as increased risk factors 
(nervous system).  

Regarding chronic pain conditions, patients with a history of abnormal spinal curvatures (which can produce low back pain and neuropathy), idiopathic peripheral neuropathies, and headaches (musculoskeletal and nervous systems) are at increased risk for addiction.  These are not surprising as they 
are notoriously difficult to treat using 
non-opioid therapies such as non-steroidal anti-inflammatory drugs (NSAIDs), steroids, or common procedures and surgeries (e.g.~joint replacement or local injections) \cite{Crofford2013}.  They involve pain that may be severely intense or debilitating, sometimes unpredictable or idiopathic, and often non-specific or diffuse (pain is referred, not well localized, or difficult to describe) and thus require a cocktail of prescription medications 
or invasive procedures, increasing the likelihood of exposure to opioids 
\cite{volkow2016opioid,rosenblum2008opioids,Patil2015}. The intensity, duration, and non-specificity of pain may also be a reason why digestive excretory, digestive (upper/oral), and reproductive conditions also show up as moderately strong features in Table~\ref{tab:nonlinearinterpretation}.  
These diagnoses may either directly result in acute or chronic non-somatic visceral pain (hernia, diverticulitis) 
\cite{Davis2012} or relate to conditions with chronic visceral pain (e.g.~female infertility may be secondary to 
endometriosis or pelvic inflammatory diseases). 
Opioids are widely utilized for such visceral pain conditions \cite{Gebhart2000}, although often in short duration due to adverse events. 
Notably, some minor injuries causing acute or procedure-related pain (ankle fractures or hand/wrist sprains -- musculoskeletal system) also feature prominently, as do features related to the mouth (dentofacial abnormalities and soft oral tissues -- digestive system (upper/oral)). This is also expected given that opioids are most commonly prescribed for post-surgical or intense acute pain \cite{Harbaugh2018,Brummett2017}; regarding dental procedures specifically, previous research suggests that a substantial proportion of adults are first exposed to opioids through dental procedures \cite{Schroeder2019}. 

Another expected finding was that individuals with psychological comorbidities (mental health conditions) also have high probability of belonging to the enhanced response group, with individuals participating in psychotherapy having reduced risk of addiction (psychotherapy was 
among the lowest-scored features and hence not shown in the table). Substantial research has already linked mental illness 
with opioid misuse \cite{glanz2018prediction,volkow2016opioid,rosenblum2008opioids}. 
Although adjustment reaction (psychology) appears with a lower score in Table~\ref{tab:nonlinearinterpretation}, it encompasses reactions to trauma, episodic emotional disorders, and chronic anxiety that have been shown to be comorbid with many of the chronic diagnoses and pain conditions discussed above \cite{glanz2018prediction,volkow2016opioid,rosenblum2008opioids} and have also been linked with increased opioid dosages \cite{helmerhorst2014risk}.  Similarly, nondependent abuse of drugs also has a lower score but 
it is well known that opioid dependence and addiction are associated with 
polysubstance use and abuse \cite{soyka2015alcohol,pergolizzi2012dynamic}.

A small subset of features with high scores were more challenging to interpret, likely because they pick up on subtle relationships between existing clinical variables or hidden variables.  For example, electrolyte imbalances (nutrition) occur often and in many situations, making it difficult to speculate on why they were a top feature for risk propensity. Likewise, skin infections and abscesses (integumentary system) 
also are common and non-specific.  However, it is possible that these features are secondary symptoms of 
important risk features. For instance, electrolyte imbalances are commonly seen in alcoholism and substance use disorders \cite{palmer2017electrolyte}, as are skin infections and abscesses \cite{springer2018integrating}. 

Based on this initial overview, the SME judged the majority of the identified features to be scientifically meaningful with potential clinical utility for future prescribing guidelines.  In summary, acute or chronic conditions that put patients at  increased risk for initial exposure to opioids, via acute procedures or comorbid prolonged intense pain, 
both increased a patient's addiction likelihood.  Co-morbid mental health disorders, particularly those related to stress or trauma and 
substance abuse also put individuals at greater risk for future opioid addiction.

\section{Conclusion}

We presented a Heterogeneous Effect Mixture Model (\method{}) for inferring subgroups of individuals that exhibit an enhanced effect caused by treatment. Our work contrasts with existing heterogeneous effect estimation methods as we learn interpretable subgroups using soft assignments while retaining expressiveness in the model. The latter is attributed to the 
capabilities of neural networks, used here to adjust for confounding. We evaluated the performance of \method{} on a synthetic dataset, the semi-synthetic IHDP dataset, and a large real-world healthcare claims dataset (\dsopioid{}). 

We additionally conducted qualitative analysis of the results obtained by \method{} on the \dsopioid{} dataset.
Some of our findings are in accordance with existing CDC opioid prescribing guidelines. However, our interpretations are preliminary and future analyses are needed to better understand these features and their relationships. 
A longer-term goal is to translate such insights into a policy white paper on data-driven, causally-valid opioid prescribing guidelines. 

\section*{Acknowledgments}
This work was conducted under the auspices of the IBM Science for Social Good initiative.  The authors thank Ching-Hua Chen, Fredrik Johansson, Aleksandra Mojsilovi\'c, Peder Olsen, Jinghe Zhang, and colleagues at IBM Watson Health for assistance.





\bibliographystyle{ACM-Reference-Format}
\bibliography{ref}

\appendix

\section{Identifiability}
\label{appx:identifiability}

\begin{manualtheorem}{1}[Identifiability]
Under the Directed Acyclic Graph in Figure. \ref{fig:plate}, $$p(Y|\textbf{do}(T=t), X) = \int_Z p(Y|X,Z,T=t) p(Z|X)$$. 
\end{manualtheorem}
\vspace{-2em}
\noindent Proof.
\begin{align*}
p(Y|\textbf{do}(T=t), X) &= \int_Z p(Y|\textbf{do}(T=t), Z, X)p(Z|\textbf{do}(T=t), X)  \\
&(\text{conditioning on and marginalizing out } Z)  \\
\text{Now,  } p(Y|\textbf{do}(T=t), Z, X) &= p(Y|T=t, Z, X)  \\
\text{and,  } p(Z|\textbf{do}(T=t), Z ) &= p(Z|T=t, X )  \\
\text{(From }& \text{\cite{pearl2009causality}'s Backdoor Adjustment Formula)}  \\
p(Y|\textbf{do}(T=t), X) &= \int_Z p(Y|\textbf{do}(T=t), Z, X)p(Z|T=t, X)  \\
&\text{(But, under  the DAG assumptions}, Z \independent T  |  X)  \\
\text{Thus, } p(Y|\textbf{do}(T=t), X) &= \int_Z p(Y|X,Z,T=t) p(Z|X) \qquad \qquad \quad \blacksquare  
\end{align*} 

\section{Parameter Inference with EM}
In this section we provide an alternate approach to perform parameter inference using Expectation Maximization and compare it to the proposed ELBO optimization.
\subsection{Inference}

The complete-data log-likelihood used in EM is given by 
\begin{equation}
	\mathcal{L}_c(\mathbf{\Theta}, \mathcal{D}) =  \sum_{i=1}^{N}\sum_{k=1}^{K}\mathbbm{1}\{z_i=k\} \ln(P^{m}_k(\mathbf{x}_i) P^{t}_{k}(y_i)),
\label{eqn:completell}
\end{equation}
here, $P^{m}_k(\mathbf{x}_i) = p(z_i=k|\mathbf{x}_i)$ 
,$P^{t}_k(\mathbf{x}_i) = p(y_i|\mathbf{x}_i, t_i, z_i=k)$ and $\mathbbm{1}$ is the indicator function.\\

\noindent \textbf{E-Step}

\noindent As is standard in EM, let us define $Q(\mathbf{\Theta}, \mathbf{\Theta}^{l})$ as the expected value of the complete-data log-likelihood \eqref{eqn:completell} with respect to the conditional distribution of the latent variable given the current parameters $\mathbf{\Theta}^l$:
\begin{displaymath}
	Q(\mathbf{\Theta}, \mathbf{\Theta}^{l}) = \mathbb{E}\left[\mathcal{L}_c(\mathbf{\Theta}, \mathcal{D}) \given \{y_i, \mathbf{x}_i, t_i\}_{i=1}^N; \mathbf{\Theta}^l \right].
\end{displaymath}
Since the only quantity in \eqref{eqn:completell} that depends explicitly on $z_i$ is the indicator $\mathbbm{1}\{z_i=k\}$, we can compute $Q$ by replacing these indicators with the posterior probability 
of $z_i=k$:
\begin{align}
&\mathbb{E}[\mathbbm{1}\{z_i=k\}] \equiv h^{(k)}_i = p(z_i = k | y_i, \mathbf{x}_i, t_i; \mathbf{\Theta}^l)\nonumber\\
&=\frac{p(y_i | z_i=k, \mathbf{x}_i, t_i; \mathbf{\Theta}^l) p(z_i=k| \mathbf{x}_i, \mathbf{\Theta}^l)}{p(y_i | \mathbf{x}_i, t_i, \mathbf{\Theta}^{l})}\nonumber\\
&=\frac{p(y_i | z_i=k, \mathbf{x}_i, t_i; \mathbf{\Theta}^l)}{p(y_i | \mathbf{x}_i, t_i, \mathbf{\Theta}^{l})} \frac{p(\mathbf{x}_i|z_i=k; \mathbf{\Theta}^l) p(z_i=k; \mathbf{\Theta}^l)}
{ p(\mathbf{x}_i; \mathbf{\Theta}^{l})}.\label{eqn:h_i^k}
\end{align}
The terms in the numerator can be evaluated using \eqref{eqn:Z}--\eqref{eqn:Xdisc}, \eqref{eqn:outcome} from the model. 
The terms in the denominator 
are normalization constants that ensure the probabilities sum to one.\\

\noindent \textbf{M-Step}

\noindent We use a gradient ascent method in the M-step to maximize $Q$ with respect to the parameters of the model:
\begin{multline}\label{eqn:Mstep}
	\mathbf{\Theta}^{l+1} =\\ \arg\max_{\mathbf{\Theta}} \left(\sum_{i=1}^{N}\sum_{k=1}^{K}h_i^{(k)} \ln[P^{m}_k(\mathbf{x}_i) P^{t}_{k}(y_i)] - \lambda \Omega(\boldsymbol{\pi}) \right).
\end{multline}
The posterior probabilities $h_i^{(k)}$ are fixed from the E-step. Using Bayes' rule as in \eqref{eqn:h_i^k}, $P^m_k(\mathbf{x}_i) = p(z_i=k\given \mathbf{x}_i)$ can be expressed in terms of model parameters $\boldsymbol{\mu}_k$, $\Sigma_k$, $\boldsymbol{\pi}_k$ defined by \eqref{eqn:Z}--\eqref{eqn:Xdisc}. Similarly, $P^t_k(y_i)$ depends on parameters $\mathbf{w}$ and $\gamma_k$ according to the outcome model \eqref{eqn:outcome}. The use of gradient ascent allows for 
any differentiable nonlinear function $f(\cdot)$ in \eqref{eqn:outcome}. Lastly, the log-prior term $-\lambda \Omega(\boldsymbol{\pi})$ favors sparsity in $\boldsymbol{\pi}$ according to either \eqref{eqn:Laplace} or \eqref{eqn:groupSparse}, where $\lambda$ is a parameter controlling the strength of the prior.

Instead of computing $Q$ over the entire dataset, we sample a mini-batch from the dataset and perform the E-step and M-step over just the mini-batch in each iteration. We observe that this mini-batch procedure is faster than regular EM over the entire dataset.

\subsection{Comparison to ELBO Optimization}

In order to compare the performance of EM vis-à-vis the variational inference-motivated ELBO optimization, we compare the train and test negative log-likelihood for both approaches on 100 realizations of the \dsihdp{} dataset, with the number of latent components, $K=3$ and stochastic gradient descent learning rate of $1\times 10^{-3}$. We then average over the resulting 100 curves. \begin{figure}[!htbp]
    \centering
    \includegraphics[width=\linewidth]{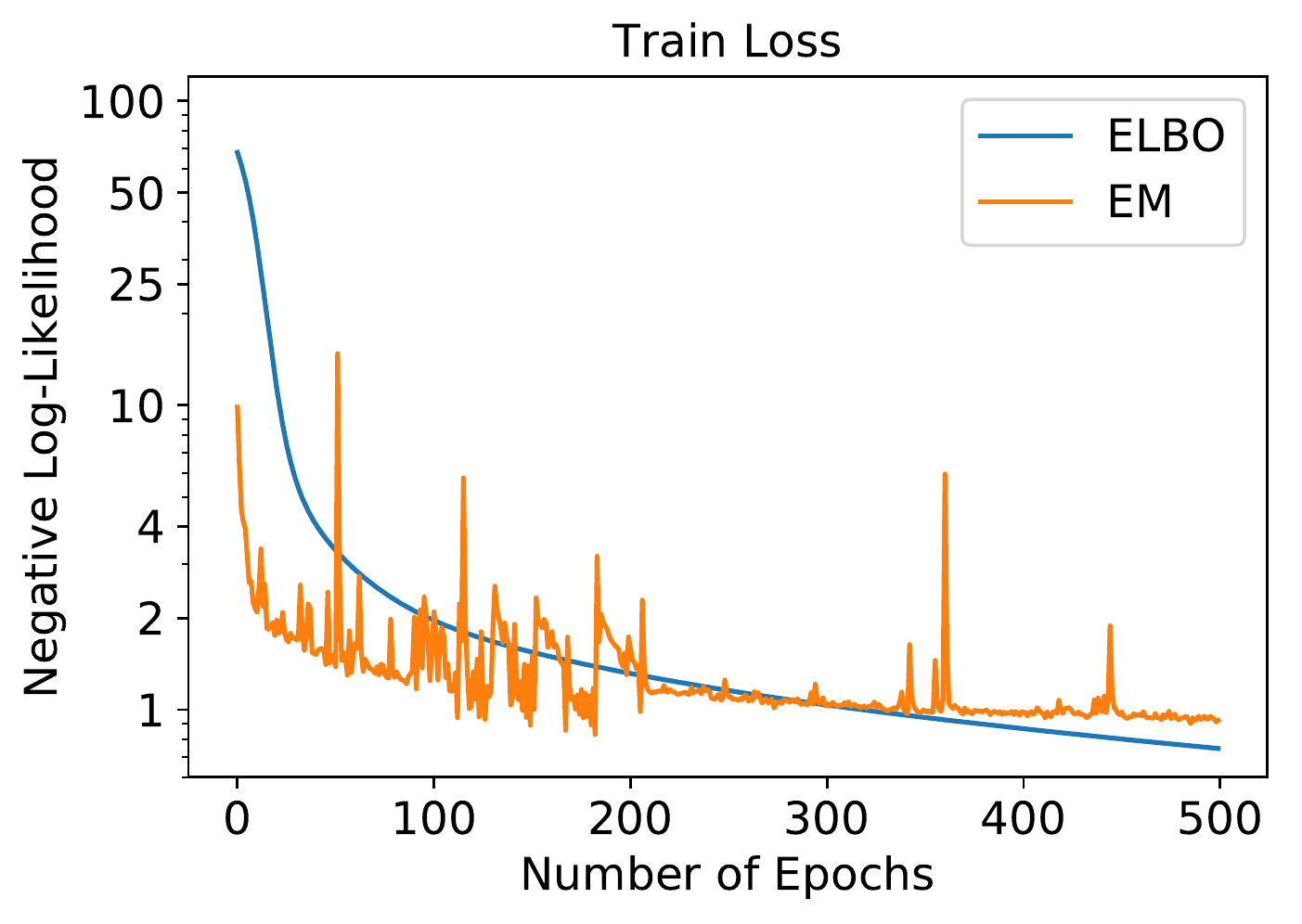}
    \includegraphics[width=\linewidth]{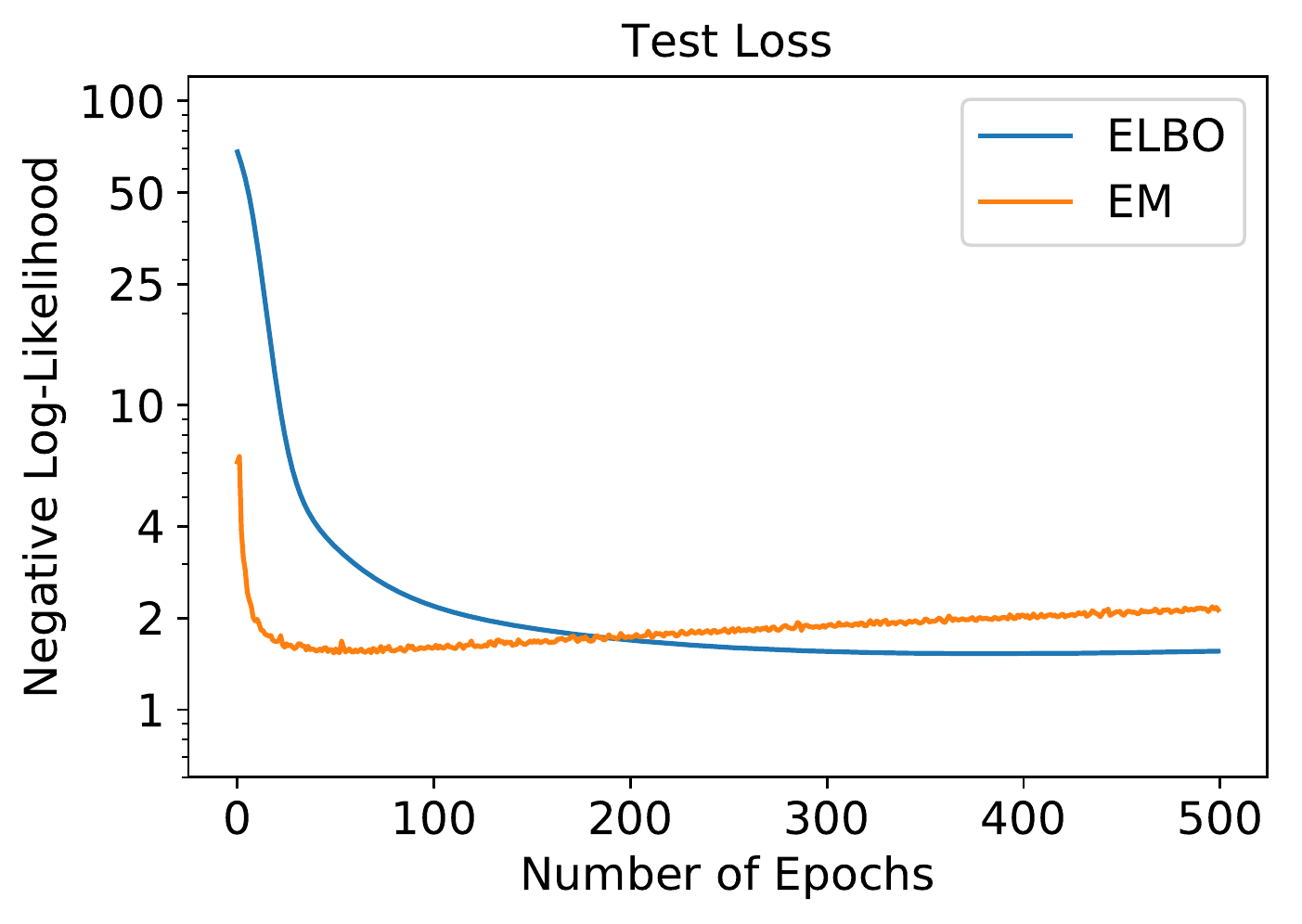}
    \caption{The negative log-likelihood (NLL) versus the number of optimization epochs for EM and ELBO. Notice how the Test NLL continues to decrease for ELBO vs.\ EM, suggesting the ELBO approach is less sensitive to overfitting.}
    \label{fig:testem}
\end{figure}
%
Figure \ref{fig:testem} presents the results; it is clear from the figure that the ELBO approach has less tendency to overfit and results in an overall better fit compared to the EM approach. This motivates our choice to directly optimize the ELBO.

\section{Parameter Initialization}
\label{sec:inference:init}

Gradient based optimization strategies can be subject 
to local minima and hence their performance is dependent on parameter initialization. To initialize the model with `good' values ensuring better convergence, we set the mean for each component, $\boldsymbol{\mu}_k$ and $\boldsymbol{\pi}_k$, equal to the sample mean of the entire data, i.e.\ $\boldsymbol{\mu}_k^0 = \frac{1}{N}\sum_i \mathbf{x}_{\text{cont},i}$, $\boldsymbol{\pi}_k^0 = \frac{1}{N}\sum_i \mathbf{x}_{\text{disc},i}$, and the covariance of every component to $\Sigma_k^0 = \diag(\bm{\sigma})$, where $\bm{\sigma}$ is a vector consisting of the sample variances of the continuous covariates $\mathbf{X}_\text{cont}$. 
We pre-train the parameters $\mathbf{w}_t$ in the outcome model \eqref{eqn:outcome} 
using standard cross-entropy loss without the subgroup and treatment assignment term $\gamma_k t$. Finally, we initialize the treatment coefficients, $\gamma_k$, randomly with positive values for all $k$.

\section{Model Fitting}
\label{sec:expt:fit}

Our implementation of \method{} has two free parameters, the number of groups $K$ and the strength of the sparsity prior, $\lambda$. 

For the \dsopioid{} dataset we divide the dataset into 3 parts with 70\% as \textsc{train} for model training, 10\% as \textsc{dev} for parameter tuning, and 20\% as \textsc{test} for evaluation. The partition is done so that the joint distribution of outcome and treatment is approximately the same in the 3 sets: $p_{\textsc{TRAIN}}(Y, T) \approx p_{\textsc{TEST}}(Y, T) \approx p_{\textsc{DEV}}(Y, T)$. For \dsihdp{} we use the stnadard 80/20 \textsc{train}/\textsc{test} split as is popular in literature.

We perform a grid search over $K \in \{2,3,4\}$ and $\lambda \in \{0, 10^{-3}, 10^{-2}, 10^{-1}\}$. 
%
For each $(K, \lambda)$ pair, we perform 5 runs with randomly initialized values of the treatment coefficients $\gamma_k$. All other parameters are initialized as described in Section~\ref{sec:inference:init}.
%
For Adam, we use a step size of $10^{-4}$ and mini-batch sizes of 10, 20 \& 1000 for \dssyn{}, \dsihdp{}, and \dsopioid{} respectively, and stop parameter update if the ELBO on the \textsc{dev} is lower at the end of an epoch.  We also search over the space of models where the outcome and counterfactual have the same or different parameterisation based on treatment assignment.  From all the $(K, \lambda)$ pairs and random initializations above, we select the model that has the best performance on the \textsc{dev} set in predicting the outcome $y_i$, in terms of the Area Under the Receiver Operating Characteristic (AU-ROC).

For the \dssyn{} dataset, we simply set $K=2$ and $\lambda=0$. In this case there is no need for a \textsc{dev} set and the data is split 50/50 between \textsc{train} and \textsc{test}. 

\begin{figure}[!htbp]

  \begin{minipage}{0.5\linewidth}
  \centering
  \includegraphics[width=\linewidth]{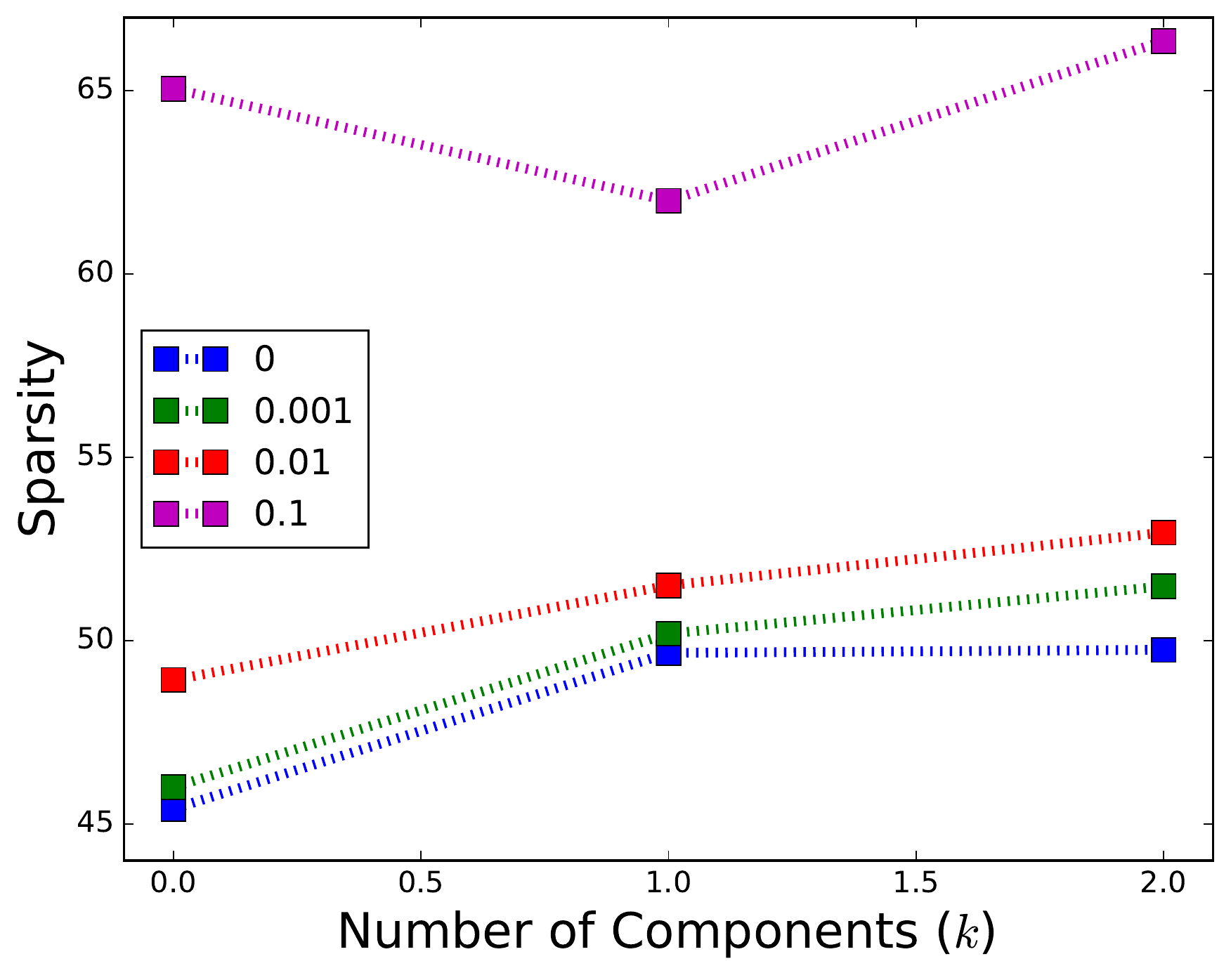}
    \end{minipage}%
    \begin{minipage}{0.5\linewidth}
      \centering
  \includegraphics[width=\linewidth]{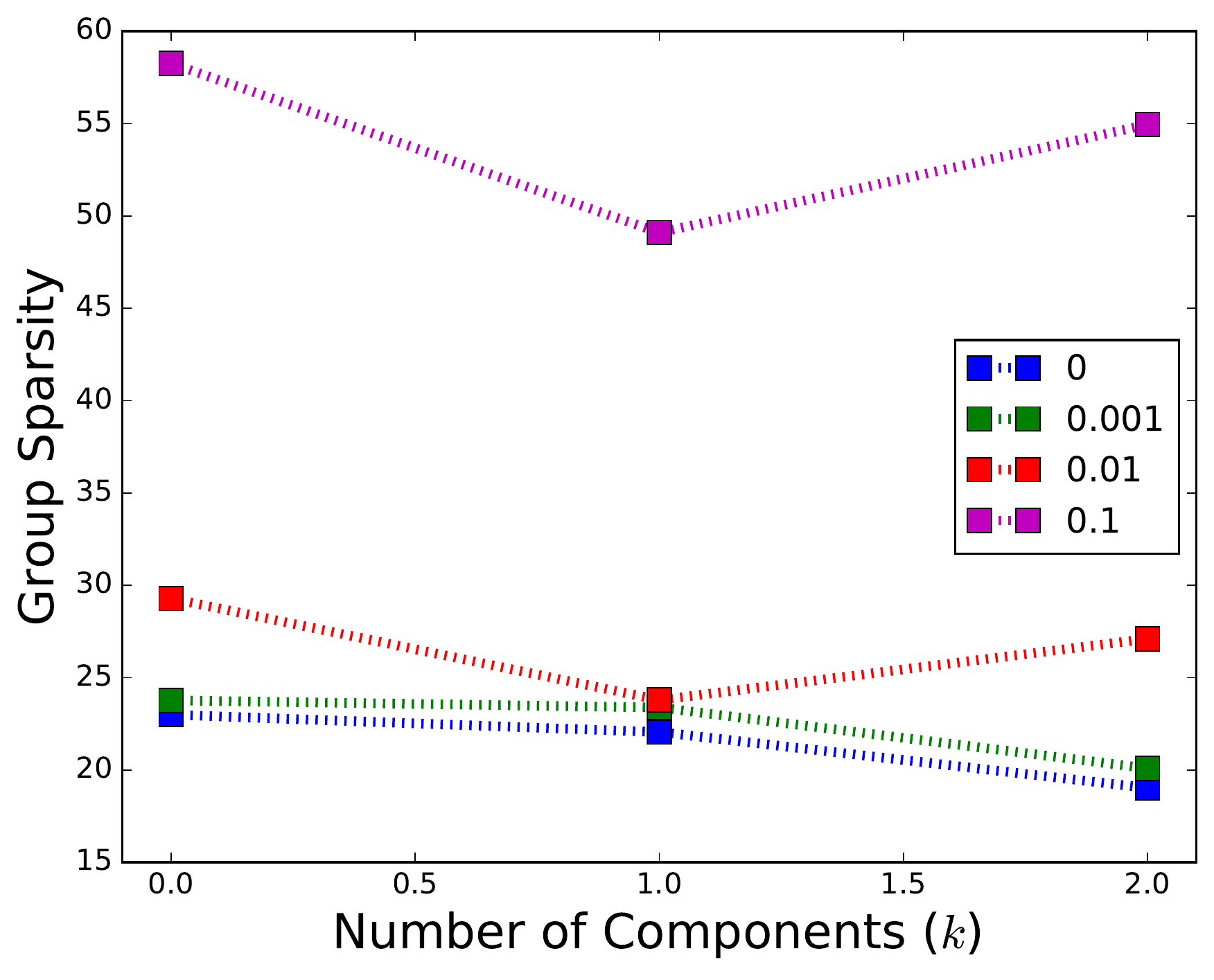}

    \end{minipage}

  \caption{Effect of the prior strength $\lambda$ for Laplace ($\ell_1$, left) and group $\ell_{1,2}$ (right) priors and different values of $K$.}
    \label{fig:groupSparse}
\end{figure}

Figure~\ref{fig:groupSparse} shows the percentage of parameters $\pi_{jk}$ equal to zero (labeled ``Sparsity'' and ``Group Sparsity'' in the plots) across all groups $k$ for different values of $(K, \lambda)$ and averaged over random initializations. It is clear that larger values of the prior strength parameter $\lambda$ result in sparser solutions. 

\end{document}